\documentclass[twoside,11pt,preprint]{article}

\usepackage{blindtext}
\usepackage{tabularx}
\usepackage{multirow}
\usepackage{booktabs}
\usepackage{array}
\usepackage{amsmath}
\usepackage{algorithm}
\usepackage{algpseudocode}
\usepackage{placeins}  

\usepackage{dmlr2e}
\usepackage{tikz}

\definecolor{GoldenTainoi}{HTML}{FFCA57}
\definecolor{AthensGray}{HTML}{DBDCE1}

\usepackage[final]{changes}  

\DeclareMathOperator*{\argmax}{arg\,max}

\usepackage{lastpage}

\makeatletter
\def\@maketitle{\vbox{\hsize\textwidth
 \linewidth\hsize \vskip \beforetitskip
 {\begin{center} \Large\bf \@title \par \end{center}} \vskip \aftertitskip
 {\def\and{\unskip\enspace{\rm and}\enspace}%
  \def\addr{\small\it}%
  \def\email{\hfill\small\sc}%
  \def\name{\normalsize\bf}%
  \def\AND{\@endauthor\rm\hss \vskip \interauthorskip \@startauthor}
  \@startauthor \@author \@endauthor}
  \vskip \aftermaketitskip
  \if@preprint
  \else
  \noindent \@starteditor \@editor \@endeditor
  \vskip \aftermaketitskip
  \fi
}}
\makeatother

\ShortHeadings{DeXposure: Decentralised Credit Exposure Dataset and Benchmarks}{Wu et al.}
\firstpageno{1}

\renewcommand{\thefootnote}{\fnsymbol{footnote}}

\begin{document}

\title{DeXposure: A Dataset and Benchmarks for Inter-protocol Credit Exposure in Decentralized Financial Networks} 

\author{\name Wenbin Wu\footnotemark[1] \email w.wu@jbs.cam.ac.uk \\
       \name Kejiang Qian\footnotemark[2] \email k.qian-8@sms.ed.ac.uk \\
       \name Alexis Lui\footnotemark[1] \email a.lui@jbs.cam.ac.uk \\
       \name Christopher Jack\footnotemark[1] \email c.jack@jbs.cam.ac.uk \\
       \name Yue Wu\footnotemark[1] \email y.wu@jbs.cam.ac.uk \\
       \name Peter McBurney\footnotemark[3] \email peter.mcburney@kcl.ac.uk \\
       \name Fengxiang He\footnotemark[2] \email f.he@ed.ac.uk \\
       \name Bryan Zhang\footnotemark[1] \email b.zhang@jbs.cam.ac.uk
       }

\editor{}

\maketitle

\footnotetext[1]{Cambridge Centre for Alternative Finance, Judge Business School, University of Cambridge}
\footnotetext[2]{School of Informatics, University of Edinburgh}
\footnotetext[3]{Department of Informatics, King's College London}

\renewcommand{\thefootnote}{\arabic{footnote}}
\setcounter{footnote}{0}

\begin{abstract}

We curate the {DeXposure} dataset, the first large-scale dataset for {inter-protocol credit exposure in decentralized financial networks}, 
covering global markets of 43.7 million entries across 4.3 thousand protocols, 602 blockchains, and 24.3 thousand tokens, from 2020 to 2025. A new measure, value-linked credit exposure between protocols, is defined as the inferred financial dependency relationships derived from changes in Total Value Locked (TVL). We develop a token-to-protocol model using DefiLlama metadata to infer inter-protocol credit exposure from the token's stock dynamics, as reported by the protocols. 

Based on the curated dataset, we develop three benchmarks for machine learning research with financial applications: (1) graph clustering for global network measurement, tracking the structural evolution of credit exposure networks, (2) vector autoregression for sector-level credit exposure dynamics during major shocks (Terra and FTX), and (3) temporal graph neural networks for dynamic link prediction on temporal graphs. From the analysis, we observe (1) a rapid growth of network volume, (2) a trend of concentration to key protocols, (3) a decline of network density (the ratio of actual connections to possible connections), and (4) distinct shock propagation across sectors, such as lending platforms, trading exchanges, and asset management protocols.

The DeXposure dataset and code have been released publicly. We envision they will help with research and practice in machine learning as well as financial risk monitoring, policy analysis, DeFi market modeling, amongst others. The dataset also contributes to machine learning research by offering benchmarks for graph clustering, vector autoregression, and temporal graph analysis. 

\noindent\textbf{Data and code:} \url{https://github.com/dthinkr/DeXposure}
    
\noindent\textbf{Visualisation:} \url{https://ccaf.io/defi/ecosystem-map/visualisation/graph}
\end{abstract}


\newpage
\tableofcontents

\newpage
\section{Introduction}

Decentralized Finance (DeFi) has emerged as a financial innovation that leverages blockchain technologies \citep{zhang2023machine} to provide autonomous financial services \citep{Werner2022-ce}. As DeFi grows in both complexity and scale \citep{fsb2023defi, aramonte2021defi}, understanding its structure, dynamics, and risk becomes increasingly essential for stakeholders, including policymakers, central bankers, and regulators. Existing studies have significantly advanced our understanding via Total Value Locked (TVL), which refers to the aggregate on-chain value of cryptoassets deposited in the smart contracts of a given protocol, blockchain, or the DeFi ecosystem as a whole \citep{saggese2025towards}. 

However, these findings are often theoretical and scenario-based \citep{esrb2023crypto_defi}, or limited to specific measurement issues without providing a system-wide quantitative map of inter-protocol credit exposures \citep{Luo2024-kj}.
This highlights a critical need for a large-scale, real-world dataset modelling TVL information with network modeling to enable macro-level, quantitative analysis for the DeFi ecosystem. 

This paper aims to address this gap by constructing DeXposure, the first large-scale dataset for inter-protocol credit exposure in decentralized financial networks, encompassing thousands of protocols that represent the majority of DeFi. Here, A DeFi protocol is a set of programs implemented through smart contracts \citep{auer2024tech_defi} on a blockchain. These programs generate digital claims that represent credit derived from underlying assets, which ultimately originates from the native tokens of the respective blockchains \citep{auer2025crypto_functions}. 

The collected DeXposure dataset comprises 43.7 million entries covering 4.3 thousand protocols, 602 blockchains, and 24.3 thousand unique tokens, spanning 2020 to 2025, providing unprecedented scale and temporal coverage for analysis in the field. We introduce a new concept \emph{value-linked credit exposure} that refers to the inferred financial dependency relationships between protocols derived from changes in their TVL patterns. Credit exposure emerges when tokens generated by one protocol (the issuing protocol) are held or locked within other protocols. By constructing this inter-protocol credit exposure network, we aim to reveal broader patterns and dynamics of exposure relationships within the DeFi ecosystem that were previously difficult to discern. 

This approach overcomes the limitations of existing studies that often focus on a small number of protocols or rely solely on transactional metrics \citep{kitzler2022defi_compositions, badev2023interconnected_defi}. It allows a more comprehensive view of exposure changes within the ecosystem, tracking how financial dependencies between protocols evolve. 

We develop three benchmarks for machine learning tasks with financial applications: (1) graph clustering for measuring the global credit exposure network \citep{chanlau2018systemic_communities}, (2) vector autoregression for analyzing network dynamics during major market shocks (such as the collapses of Terra \citep{lee2023terra_luna_case} and FTX \citep{conlon2024ftx_contagion}), and (3) temporal graph neural networks \citep{rossi2020tgn} for predicting dynamic links \citep{Dileo2024-ep}. Our findings can provide implications for policymakers, central bankers, and regulators by revealing exposure dynamics, market structure evolution, and systemic risk factors. These results can also enhance the capacity to monitor and predict the dynamics of the DeFi market. 

To support reproducibility, we publicly release our dataset and code at \url{https://github.com/dthinkr/DeXposure}. We also present our results through an interactive DeFi digital tool \url{https://ccaf.io/defi/ecosystem-map/visualisation/graph}.
\FloatBarrier
\section{Related Works}

\
This section provides an overview of DeFi protocols, the TVL metric, and recent research in the field, setting the context for our study.

\subsection{DeFi Protocols}

Unlike traditional finance, where the International Organization for Standardization (ISO) provides a comprehensive framework for classifying financial instruments \citep{International-Standard-Organization2021-xw}, no such standardized taxonomy exists for DeFi. This lack of standardization complicates analysis. 

\citet{Zetzsche2020-pv} broadly classify DeFi applications into four categories: (1) deposit taking and lending, (2) trading and investments, (3) insurance, and (4) auxiliary services. Examples include lending protocols such as Aave and Compound, decentralized exchanges such as Uniswap, and yield aggregators such as Yearn Finance \citep{Schar2021-defi, auer2024tech_defi}. 

\citet{Werner2022-ce} provide a more granular classification, identifying seven main categories of DeFi protocols: (1) decentralized exchanges, (2) lending platforms, (3) derivatives, (4) asset management, (5) insurance, (6) payment channels, and (7) stablecoins. Their systematization maps these categories to prominent protocols such as Uniswap (DEX), Compound (lending), Synthetix (derivatives), and DAI (stablecoin) \citep{Schar2021-defi, auer2024tech_defi}. 

\citet{Xu2021-bh} provide a systematic analysis of Decentralized Exchanges (DEXs) with Automated Market Maker (AMM) protocols, highlight the transition from traditional banking to DeFi lending \citep{Xu2022-ni}, and provide a comprehensive survey of business models across different types of DeFi protocols \citep{Xu2023-pf}. However, protocol-based classifications often fall short for multi-utility protocols, where one protocol is associated with multiple categories. Understanding each protocol's underlying assets and liabilities would pave the way for instrument-based classification, which aligns with established taxonomies \citep{International-Standard-Organization2021-xw}, though this remains challenging due to data heterogeneity.

DefiLlama, a DeFi analytics platform, currently categorizes protocols into dozens of categories \citep{DefiLlama2024-zw}. While this granularity is useful for detailed analysis, it can complicate broader ecosystem-level insights. In this work, we map these numerous DefiLlama categories into a set of broader, more concisely defined groups, which enhances clarity. Table \ref{tab:category_mapping} presents our categorization scheme.

\begin{table}[ht!]
\centering
\footnotesize
\caption{Mapping of DefiLlama Protocol Categories to Broader Analytical Groups.}
\label{tab:category_mapping}
\begin{tabularx}{\columnwidth}{>{\raggedright\arraybackslash}p{4.5cm}X}
\toprule
\textbf{Broad Category} & \textbf{DefiLlama Categories} \\
\midrule
Asset Management & Algo-Stables, Decentralized Stablecoin, Liquid Staking, Liquidity manager, Reserve Currency, Synthetics, Yield, Yield Aggregator \\
\midrule
Trading \& Exchanges & Bridge, CEX, DEX Aggregator, Cross Chain, Dexes, Derivatives, Options, Options Vault, NFT Marketplace \\
\midrule
Lending, Borrowing \& Real World Assets & CDP, Lending, Leveraged Farming, NFT Lending, RWA Lending, Uncollateralized Lending, RWA, Liquidity Restaking, Restaking \\
\midrule
Infrastructure, Services \& Financial Products & Chain, Infrastructure, Oracle, Payments, Services, Farm, Gaming, Indexes, Insurance, Launchpad, Prediction Market, Staking Pool, Wallets \\
\midrule
Privacy \& Security & Privacy \\
\midrule
Others & SoFi \\
\bottomrule
\end{tabularx}
\end{table}

\subsection{Credit Exposure in DeFi}

In traditional finance, credit exposure represents the potential loss a lender faces if a borrower defaults \citep{BCBS-counterparty}. In DeFi, credit exposure arises from token-mediated dependencies: when one protocol holds tokens issued by another, it becomes exposed to that protocol's credit risk. Unlike traditional finance, where exposure is bilateral and explicitly contracted, DeFi exposure relationships form organically through protocol composability and user actions.

The token-based nature of DeFi creates complex networks of interdependency. A protocol that issues tokens establishes potential exposure pathways whenever those tokens are adopted as collateral, liquidity, or reserves by other protocols. These exposure relationships can propagate contagion during market stress, as the failure of one protocol can affect all protocols that hold its tokens. Recent empirical studies have examined credit exposure at the protocol level through liquidation mechanisms \citep{Qin2021-liquidations} and lending risk analysis \citep{Doerr2025-defi-lending}, providing evidence of the materiality of these exposure relationships.

Existing research has primarily focused on static network snapshots or small-scale analyses of individual lending protocols, based on transaction-level data. As DeFi grows in scale and interconnection with traditional finance, there is a critical need to track credit exposure dynamics across the entire DeFi ecosystem over extended time periods for regulatory monitoring, systemic risk oversight, and policy formulation. Understanding how exposure concentrations shift during market events, how new exposure pathways emerge, and how the network's topology evolves provides essential insights for financial stability assessment, early warning systems, and the development of appropriate regulatory frameworks.

\subsection{Total Value Locked (TVL)}

Total Value Locked (TVL) is a fundamental metric in decentralized finance that measures the total USD value of digital assets deposited and held within a DeFi protocol at any given time \citep{DefiPulse2021-ai}. Developed by DeFi Pulse \citep{DefiPulse2021-ai} and popularized by DefiLlama \citep{DefiLlama2021-je}, TVL serves as a key indicator of a protocol's size, popularity, and market significance. For example, if users have deposited \$100 million worth of various cryptocurrencies (such as Bitcoin \citep{Nakamoto2008-cq}, Ethereum \citep{Buterin2014-ethereum}, or fiat-referenced stablecoins \citep{Lyons2020-ie}) into a lending protocol, that protocol's TVL would be \$100 million. As users deposit or withdraw assets, the TVL fluctuates accordingly, providing insights into capital flows and protocol adoption. The metric quantifies the token assets held within a DeFi protocol and computes their currency value based on associated token prices. The calculation of TVL faces challenges due to the heterogeneous nature of DeFi protocols, which lack a unified accounting standard.

Additionally, DeFi protocols function primarily as credit instruments, leading to issues of double counting or \textit{rehypothecation}, similar to the money multiplier in banking \citep{McLeay2014-fo}. While recent studies \citep{Luo2024-kj} have proposed methodologies to mitigate double counting, our perspective is that without a clear outline of credit and liquidity risks, the concept of double counting remains ambiguous, as credit inherently involves multiple claims on the same underlying assets. Thus, identifying and excluding double counting based on a token's stage in circulation might overlook important market dynamics, specifically leverage and interconnectedness. Furthermore, confusion over a protocol's assets and liabilities complicates TVL calculations, as some protocols compute TVL using liability tokens rather than asset tokens \citep{DefiLlama2024-zw}, making it challenging to assess protocol solvency.

Despite these challenges, TVL remains a comprehensive metric that captures the scope of the DeFi landscape through its crowdsourced data from various protocol development teams. As of 2025, data providers such as DefiLlama track TVL for over 2,000 DeFi protocols across hundreds of blockchains \citep{DefiLlama2024-zw, Stepanova2021-uo}, providing a rich resource for analyzing the asset side of DeFi balance sheets.

\subsection{Network Analysis of DeFi}

Research on TVL, DeFi network analysis has attracted increased attention in recent years.

\citet{Luo2024-kj} introduced Total Value Redeemable (TVR) to address double counting in TVL calculations. Their analysis of 100 DeFi protocols showed TVR to be more stable than TVL during market downturns and quantified double counting through a DeFi money multiplier. \citet{Metelski2022-kf} used correlation analysis and panel data models to examine relationships between TVL and other economic indicators, finding that protocol valuations depend on various performance measures, with TVL's impact varying across metrics. \cite{Stepanova2021-uo} tracked TVL for 12 major DeFi applications over a 34-month period, documenting the rapid growth and concentration of activity in a small set of protocols.

Network analysis has provided key insights into DeFi's complex interactions. \cite{kitzler2022defi_compositions} studied compositions of 23 major DeFi protocols on Ethereum and constructed a smart-contract interaction network, showing that decentralized exchanges and lending protocols occupy central positions and that interactions concentrate in a strongly connected component, making DeFi structurally prone to contagion. \cite{Li2024-ha} employed network analysis to research systemic risk in 30 DeFi protocols, finding that protocol interconnectedness can lead to contagion effects during market stress. \cite{Alamsyah2024-mx} analyzed 5.8 million transactions of three DeFi tokens on Ethereum. Using network metrics, they quantified market size, transactions per wallet, market density, and transaction clusters. Their centrality calculations identified key wallet addresses and their market roles.

Several studies have focused on specific aspects of DeFi. \cite{Znaidi2023-jd} analyzed the Curve ecosystem over a 2-year period, highlighting its central role in DeFi. \cite{Weingartner2023-wx} examined cross-chain bridges in DeFi, emphasizing their growing importance and associated risks. \cite{Zhang2023-dt} studied the evolution of automated market makers (AMMs) in DeFi over a 3-year period, tracking changes in protocol designs and market structures.

\subsection{Dataset Landscape}

The advancement of DeFi has necessitated new empirical resources, leading to a recent surge in open-access datasets. These contributions primarily target three domains: security, market microstructure, and governance.

In the security domain, researchers have released extensive datasets to combat fraud. \citet{Sun2025} and \citet{Alhaidari2025} provide granular data on rug pulls across Ethereum, BSC, and Solana, facilitating the training of detection models. \citet{Suzuki2025} bridge on-chain and off-chain data to investigate token scams, while \citet{Carpentier2025} map the broader landscape of crypto-crime incidents.

For market microstructure and economic analysis, \citet{Chemaya2025} introduced daily transaction indices for Uniswap v3, enabling cross-chain comparisons of DEX liquidity. \citet{Chen2025} established benchmarks for yield prediction on Curve Finance. In governance, \citet{Ma2024} compiled a comprehensive corpus of smart contract audit reports to analyze voting vulnerabilities.

Despite these advancements, existing datasets are largely segmented by specific verticals (e.g., fraud, specific DEXs) or event types. There remains a scarcity of longitudinal, ecosystem-wide data that captures the \textit{interconnected balance sheets} of protocols.  
\FloatBarrier
\section{Dataset Curation Methodology}
This section outlines our approach to building a large-scale DeFi network dataset designed for machine learning applications. We transform unstructured on-chain value states into a structured temporal graph \(\{G_\tau\}\) suitable for tasks such as dynamic link prediction, graph clustering, and anomaly detection. Our methodology addresses the challenge of converting raw TVL data into a coherent network topology that preserves the temporal dependencies required for learning dynamic market behaviors.

\FloatBarrier

\subsection{Preliminaries}

We begin by formally defining the foundational concepts and data structures underlying our network construction.

\subsubsection{Data Model}

We define the data model formally. Let \( \mathcal{P} \) be the set of all protocols, \( \mathcal{C} \) the set of all chains, and \( \mathcal{X} \) the set of all tokens. Each token \( \mathcal{x} \in \mathcal{X} \) at any given time \( t \) is characterized by a tuple \( (n_t, v_t) \), where \( n_t \) represents the amount of the token, and \( v_t \) represents the USD value of the token.

The relationship between protocols, chains, and tokens is modeled using a mapping:
$\mathcal{R}: \mathcal{P} \times \mathcal{C} \rightarrow 2^{\mathcal{X}}$, where \( 2^{\mathcal{X}} \) denotes the power set of \( \mathcal{X} \), representing all possible subsets of tokens with their associated attributes. This mapping allows each protocol-chain pair to be associated with a dynamic subset of tokens. For instance, consider a protocol \( p \in \mathcal{P} \) operating on a blockchain \( c \in \mathcal{C} \). The mapping \( \mathcal{R}(p, c) \) might yield a subset of tokens \( \{\mathcal{x}_1, \mathcal{x}_2, \ldots, \mathcal{x}_k\} \subset \mathcal{X} \), where each token \( \mathcal{x}_i \) at time \( t \) is represented as:
\begin{equation}
\mathcal{x}_i(t) = (n_{t_{\mathcal{x}_i}}, v_{t_{\mathcal{x}_i}}).
\end{equation}

To analyze the global state of a token across all protocols and chains, we define the global state of token \( \mathcal{x}_i \) at time \( t \) as \( \Theta_{\mathcal{x}_i}(t) \). This global state is an aggregation of the token's amounts and values from all protocol-chain pairs where the token is present and is represented by:

\begin{equation}
\Theta_{\mathcal{x}_i}(t) = \left( \sum_{(p, c) \in \mathcal{R}^{-1}(\mathcal{x}_i)} n_{t_{\mathcal{x}_i}}^{(p,c)}, \sum_{(p, c) \in \mathcal{R}^{-1}(\mathcal{x}W_i)} v_{t_{\mathcal{x}_i}}^{(p,c)} \right),
\end{equation}

where \( \mathcal{R}^{-1}(\mathcal{x}_i) \) denotes the set of all protocol-chain pairs \( (p, c) \) that include token \( \mathcal{x}_i \), \( n_{t_{\mathcal{x}_i}}^{(p,c)} \) represents the amount of \( \mathcal{x}_i \) under protocol \( p \) and chain \( c \) at time \( t \), and \( v_{t_{\mathcal{x}_i}}^{(p,c)} \) represents the USD value of \( \mathcal{x}_i \) under the same conditions.

We define two additional types of global states: protocol-wise token state and chain-wise token state:
\begin{equation}
\label{eq:protocol-wise-token-state}
\Theta_{\mathcal{x}_i}^{p}(t) = \left( \sum_{c \in \mathcal{C} | (p, c) \in \mathcal{R}^{-1}(\mathcal{x}_i)} n_{t_{\mathcal{x}_i}}^{(p,c)}, \sum_{c \in \mathcal{C} | (p, c) \in \mathcal{R}^{-1}(\mathcal{x}_i)} v_{t_{\mathcal{x}_i}}^{(p,c)} \right),
\end{equation}

\begin{equation}
\label{eq:chain-wise-token-state}
\Theta_{\mathcal{x}_i}^{c}(t) = \left( \sum_{p \in \mathcal{P} | (p, c) \in \mathcal{R}^{-1}(\mathcal{x}_i)} n_{t_{\mathcal{x}_i}}^{(p,c)}, \sum_{p \in \mathcal{P} | (p, c) \in \mathcal{R}^{-1}(\mathcal{x}_i)} v_{t_{\mathcal{x}_i}}^{(p,c)} \right).
\end{equation}

For a given protocol \( p \) or a chain \( c \) at time \( t \), equations \ref{eq:protocol-wise-token-state} and \ref{eq:chain-wise-token-state} describe the aggregated state of token \( \mathcal{x}_i \) for both amount and value. They enable the global analysis of token behavior within specific protocols or chains. For example, we can answer questions such as: how much token \(i\) is locked within a DeFi protocol, or a blockchain at a historical time point. Such answers are derived from this data model and are documented in our published tool \citep{Cambridge-Center-for-Alternative-Finance2024-fi}. 

\subsubsection{Credit Exposure Formalization}

\begin{definition}[Credit Exposure]
Let \(X_G(q)\) denote the set of tokens that protocol \(q\) generates or issues, and let \(\mathcal{X}_p(t)\) denote the set of tokens currently held (locked) by protocol \(p\) at time \(t\). Protocol \(p\) has credit exposure to protocol \(q\) if:
\begin{equation}
\mathcal{X}_p(t) \cap X_G(q) \neq \emptyset.
\end{equation}
\end{definition}

This means tokens generated by \(q\) are currently locked in \(p\), creating a financial dependency. If tokens issued by \(q\) lose value or become illiquid, protocol \(p\) faces exposure to this risk.

\begin{figure}[ht!]
    \centering
    \begin{tikzpicture}[node distance=3cm, auto]
        \tikzstyle{entity} = [rectangle, draw, fill=GoldenTainoi, minimum width=2cm, minimum height=1cm]
        \tikzstyle{balance} = [rectangle, draw, fill=AthensGray, minimum width=3.8cm, minimum height=3cm]
    
        \node[entity] (user) at (0,2) {User};
        \node[entity] (weth) at (-4,0) {WETH Protocol};
        \node[entity] (maker) at (4,0) {MakerDAO};
        
        \node[balance] (balanceWETH) at (-4,-2.5) {
            \begin{tabular}{lr}
                \textbf{WETH Protocol} & \\
                Assets: & \\
                ETH & \$100 \\
                Liabilities: & \\
                WETH & \$100
            \end{tabular}
        };
        \node[balance] (balanceMaker) at (4,-2.5) {
            \begin{tabular}{lr}
                \textbf{MakerDAO} & \\
                Assets: & \\
                WETH & \$100 \\
                Stability Buffer & \$-34 \\
                Liabilities: & \\
                DAI & \$66 \\
            \end{tabular}
        };
    
        \draw[->, thick] (user) to[bend left=30] node[above, midway] {ETH \$100} (weth);
        \draw[->, thick] (weth) to[bend left=30] node[below, midway] {WETH \$100} (user);
        \draw[->, thick] (user) to[bend right=30] node[above, midway] {WETH \$100} (maker);
        \draw[->, thick] (maker) to[bend right=30] node[below, midway] {DAI \$66} (user);

    \end{tikzpicture}
    \caption{Credit Exposure Example: WETH Wrapping and Overcollateralized DAI Generation. User wraps ETH to WETH, then uses WETH as collateral to generate DAI from MakerDAO, creating credit exposure from MakerDAO to the WETH protocol.}
    \label{fig:defi_parallel_with_provision}
\end{figure}
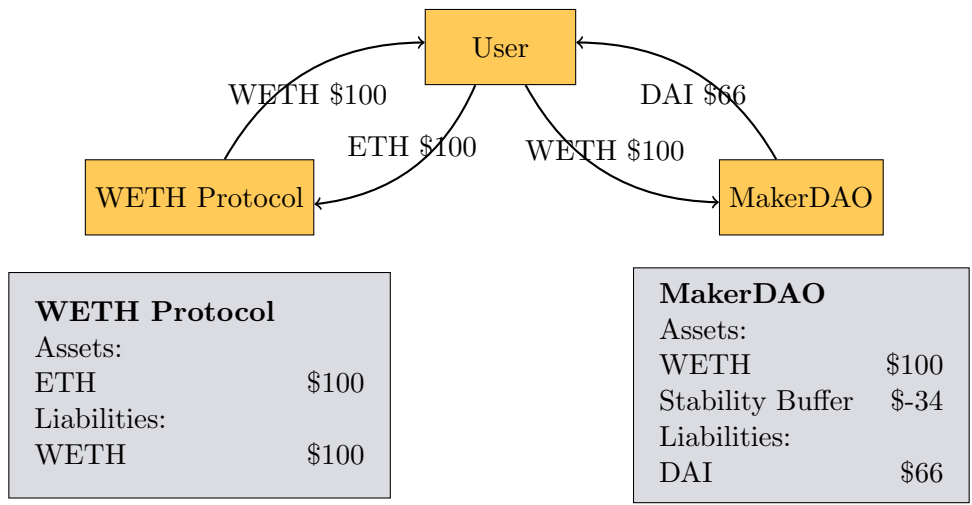

To illustrate this concept, Figure \ref{fig:defi_parallel_with_provision} shows how credit exposure emerges in DeFi. A user wraps ETH to WETH through the WETH protocol, then uses WETH as collateral in MakerDAO to generate DAI. This creates credit exposure from MakerDAO to the WETH protocol—if WETH fails, MakerDAO's collateral is at risk. The figure shows the balance sheet positions: WETH protocol holds ETH as assets with WETH as liabilities, while MakerDAO holds WETH as assets with DAI as liabilities, creating a chain of credit dependencies.

\FloatBarrier

\FloatBarrier

Our approach measures the dynamics of these credit exposure relationships over time through changes in TVL. As token holdings shift between protocols, exposure relationships strengthen, weaken, or dissolve.

\FloatBarrier

\subsection{Raw Data Collection}

\begin{figure}[htbp]
\centering
\includegraphics[width=\textwidth]{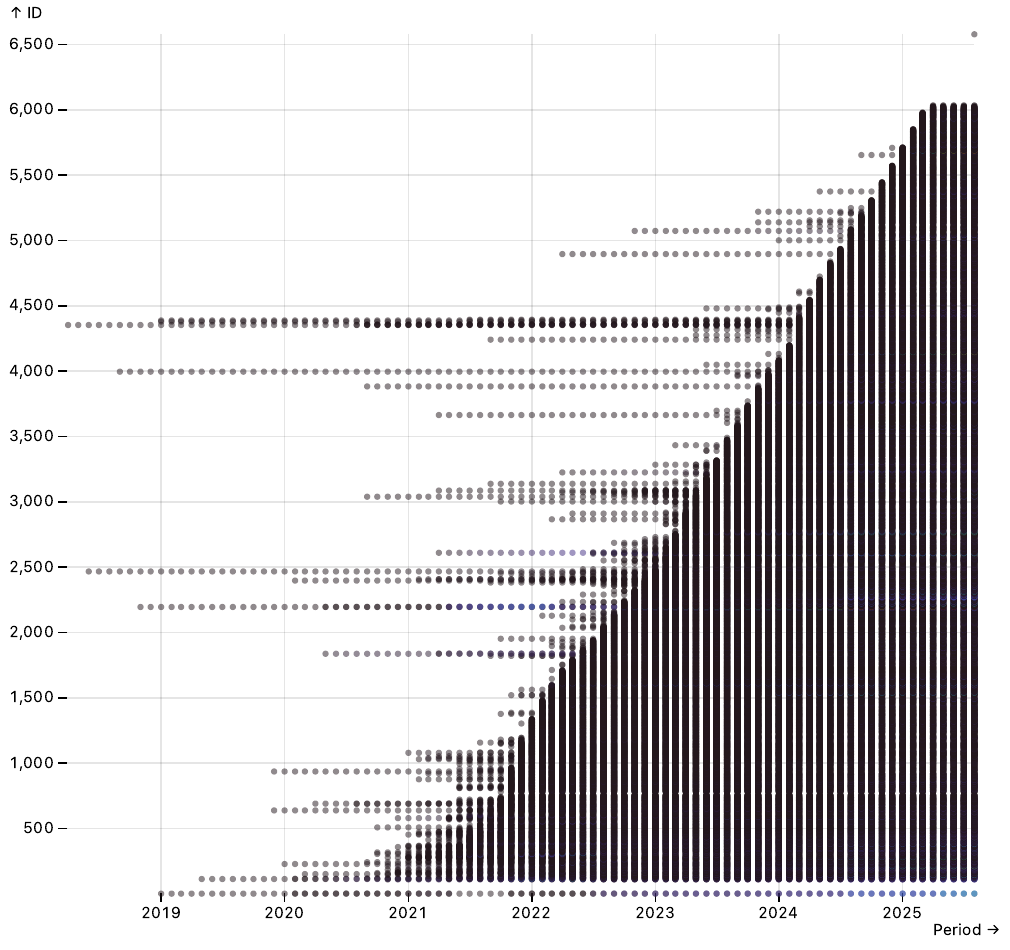}
\caption{Data Availability and Frequency Across Time. Date range from 2018 to 2024. }
\label{fig:data_availability_frequency}
\end{figure}

We retrieve our data from DefiLlama \citep{DefiLlama2021-je}. Figure \ref{fig:data_availability_frequency} offers a visual assessment of data distribution over time. The heatmap illustrates data availability and granularity for all recorded DeFi protocols, identified by numeric IDs. Each dot indicates at least one data entry per date. Color density varies with data granularity: darker shades denote higher frequencies, such as hourly updates, while lighter shades suggest less frequent updates, like daily ones.

\FloatBarrier

\FloatBarrier

\subsection{Financial Networks Construction and Representation}
\label{sec:network_model_description}

We detail the construction and analysis of our network dataset, which is designed to capture the dynamic inter-protocol credit exposure relationships between different DeFi protocols over discrete time intervals through changes in Total Value Locked. 

\FloatBarrier

\subsubsection{Handling Token Attributes}

Let \( t \) represent discrete time steps over which the network is analyzed, with \( T \) being the set of all such steps. First, define \( \mathcal{X}_p \) to represent the set of tokens associated with a protocol \( p \) at a given time \( t \). This can be expressed from the mapping \( \mathcal{R} \) which relates protocols and chains to tokens:
\begin{equation}
\mathcal{X}_p(t) = \bigcup_{c \in \mathcal{C}} \mathcal{R}(p, c).
\end{equation}

Define stock \(S_t\) as the USD value at time \(t\) of an entity, where the entity can be either a protocol \(p\) or a chain \(c\). For example, the total stock of a protocol \( p \) at a given time \( t \) is calculated by aggregating the USD value \( v_t \) of all tokens \( \mathcal{x} \) within the protocol:
\begin{equation}
S_{p,t} := \sum_{\mathcal{x} \in \mathcal{X}_p(t)} v_{t_{\mathcal{x}}}.
\end{equation}

Define time interval \(\tau := t_2 - t_1\), where \(t_1\) and \(t_2\) are consecutive time points in \(T\). The change in stock for a protocol \( p \) over the interval \( \tau \) is calculated by aggregating the changes in stock values of all tokens \( \mathcal{x} \) within the protocol over that interval. This can be represented as:
\begin{equation}
\Delta S_{p,\tau} := \sum_{\mathcal{x} \in \mathcal{X}_p(t)} \Delta v_{p,\tau}^{\mathcal{x}},
\end{equation}

where \( \Delta v_{p,\tau}^{\mathcal{x}}:= v_{p,t_2}^{\mathcal{x}} - v_{p,t_1}^{\mathcal{x}} \) is the change in stock of token \( \mathcal{x} \) at protocol \( p \) over the interval \( \tau \), and the change of token state can be described by \( \Delta \mathcal{x}_{p,\tau} = (\Delta n_{p,\tau}^{\mathcal{x}}, \Delta v_{p,\tau}^{\mathcal{x}}) \).

\FloatBarrier

\subsubsection{Token-Protocol Mapping}

Given a token \( \mathcal{x} \) in protocol \( p \), the mapping function \( \mathcal{M} \) determines the issuing protocol \( q \) that generates token \( \mathcal{x} \). The issuing protocol is the protocol that generates or manages a given token, establishing the foundation for credit exposure relationships. Using this function, we are able to map a token in protocol \( p \) to its issuing protocol \( q \). This mapping ensures that changes in token holdings in protocol \( p \) can be attributed to exposure relationships with the issuing protocol \( q \),
\begin{equation}
q = \mathcal{M}(\mathcal{x}_{p,\tau}).
\end{equation}

To ensure accurate and comprehensive mapping \(\mathcal{M}\), we design four methods that are applied in a fall-back manner:

\begin{enumerate}
    \item Tokens are directly mapped to protocols based on the metadata list sourced from DefiLlama \citep{DefiLlama2021-pz}, where token names are listed along with specific protocols. This method is straightforward and relies on source data where the relationship between tokens and protocols is already identified.

    \item In cases where direct lookup is not possible, tokens are manually mapped to protocols. This involves human intervention to categorize tokens based on expert knowledge and available documentation. We create a manual mapping for tokens with the highest average historical TVL. Table \ref{tab:tokens_to_protocols} shows a selection of the top tokens defined in this way. 

    \item For tokens that are not covered by the first two methods, we employ a more sophisticated approach involving textual analysis:
        \begin{enumerate}
            \item Textual Representation: Each token \( \mathcal{x} \) and protocol \( p \) is represented by a set of descriptive texts extracted from metadata fields, which we denote as \( \text{Texts}_{\mathcal{x}} \) and \( \text{Texts}_{p} \). 
            
            \item Vectorization: Textual data is transformed into numerical vectors using the Term Frequency-Inverse Document Frequency (TF-IDF) \citep{Qaiser2018-bn, Ramos2003-cf} method. We can therefore derive vector representations \( V_{\mathcal{x}} \) and \( V_{p} \) for each token and protocol,
            \begin{equation}
            V_{\mathcal{x}}, V_{p} = \text{TFIDF}(\text{Texts}_{\mathcal{x}}, \text{Texts}_{p}).
            \end{equation}

            \item Cosine Similarity Calculation: We calculate the similarity between the vectorized representations of tokens and protocols using the cosine similarity metric. The similarity score \( S(\mathcal{x}, p) \) is given by:
            \begin{equation}
            S(\mathcal{x}, p) = \frac{V_{\mathcal{x}} \cdot V_{p}}{\|V_{\mathcal{x}}\| \|V_{p}\|}.
            \end{equation}

            \item Thresholding and Mapping: Tokens are mapped to the protocol with the highest similarity score, provided that the score exceeds a predefined threshold \( \theta \).
            Where \( \mathcal{M}(\mathcal{x}) \) denotes the mapping of token \( \mathcal{x} \) to a protocol, this mapping is formally defined as:
            \begin{equation}
            \mathcal{M}(\mathcal{x}) = \argmax_{p \in \mathcal{P}}(S(\mathcal{x}, p)) \quad \text{if } S(\mathcal{x}, p) > \theta.
            \end{equation}
            
        \end{enumerate}

    \item If a token cannot be mapped to an existing protocol \( p \) through any of the above methods, i.e., it does not pass through the TFIDF threshold, it is categorized under a protocol with its own token name. This serves as a catch-all for tokens that do not fit into any other mapping criteria. We categorize these protocols as \textit{Primary Market Tokens}. Tokens that fall into this category are, for example, WETH, which are protocols that handle or produce a single token.  
\end{enumerate}

\begin{table}[htbp]
    \centering
    \footnotesize
    \caption{Select Content in Tokens to Protocol Mapping.}
    \label{tab:tokens_to_protocols}
    \begin{tabularx}{\columnwidth}{>{\hsize=0.4\hsize}X>{\hsize=0.6\hsize}X>{\hsize=0.4\hsize}X>{\hsize=0.6\hsize}X}
    \toprule
    \textbf{Token} & \textbf{Protocol} & \textbf{Token} & \textbf{Protocol} \\
    \midrule
    USDC & Circle & USDT & Tether \\
    WETH & WETH & DAI & MakerDAO \\
    WBTC & BitGo & ETH & Ethereum \\
    BUSD & Binance & WBNB & Binance \\
    BNB & Binance & FTM & Fantom \\
    WAVAX & Avalanche & WFTM & Fantom \\
    FRAX & Frax Finance & MIM & Abracadabra \\
    TUSD & TrustToken & UST & Terra \\
    DOT & Polkadot & ADA & Cardano \\
    WSTETH & Lido & OP & Optimism \\
    XRP & Ripple & SUSD & Synthetix \\
    RETH & Rocket Pool & DOGE & Dogecoin \\
    LUSD & Liquity & HBTC & Huobi \\
    RENBTC & Ren & MANA & Decentraland \\
    MAI & Qi Dao & STETH & Lido \\
    GRT & The Graph & LTC & Litecoin \\
    LUNA & Terra & TRX & TRON \\
    ATOM & Cosmos & AXS & Axie Infinity \\
    FEI & Fei Protocol & AGEUR & Angle \\
    DPI & Index Coop & METIS & Metis \\
    GOHM & Olympus & USDP & Paxos \\
    RAI & Reflexer & HUSD & Huobi \\
    PAXG & Paxos & ENJ & Enjin \\
    FIL & Filecoin & WXDAI & Gnosis \\
    3CRV & Curve & CEL & Celsius \\
    EURS & STASIS & BCH & Bitcoin Cash \\
    CUSD & Celo & WONE & Harmony \\
    JEUR & STASIS & GALA & Gala Games \\
    OCEAN & Ocean & DOLA & Dollar \\
    BAND & Band & NEAR & NEAR \\
    TWT & Trust Wallet & ALUSD & Alchemix \\
    WSOL & Solana & WUSDR & Tangible \\
    JPYC & JPY Coin & ANY & Anyswap \\
    SETH & Synthetix & UMA & UMA \\
    \bottomrule
    \end{tabularx}
\end{table}

\FloatBarrier

\subsubsection{Network Data Modeling}

\FloatBarrier

We describe the interactions between DeFi protocols over discrete time intervals \( \tau \) using a series of weighted directed graphs, where each graph corresponds to a snapshot of interactions within a network of DeFi protocols at discrete time intervals. Here, we focus on modeling each snapshot instead of exploring its temporal relationship. Consider each graph snapshot as \( G_{\tau} = (P_{\tau}, E_{\tau}) \). We define the weighted set of vertices as:
\begin{equation}
P=\left\{p \in \mathcal{P}: w\left(p\right) \in \mathbb{R}^{+}\right\},
\end{equation}

where $\mathcal{P}$ is the set of all possible DeFi protocols under consideration, and $w\left(p\right)$ represents the assigned weight to protocol $p$. 

The weighted edge set is defined as:
\begin{equation}
E=\left\{\left(p, q, w\left(e_{p q}\right)\right):\left(p, q\right) \in \mathcal{E}, w\left(e_{p q}\right) \in \mathbb{R}^{+}\right\},
\end{equation}
where $\mathcal{E}$ represents all possible directed interactions within the network, and $w\left(e_{p q}\right)$ quantifies the interaction strength from $p$ to $q$.

Given a time interval \( \tau := t_2 - t_1 \), the vertice weight is defined by:
\begin{equation}
w_{\tau}(p) = S_{p, \tau} := \sum_{\mathcal{x} \in \mathcal{X}_p(t_1) \cap \mathcal{X}_p(t_2)} v_{t_2, \mathcal{x}}.
\end{equation}

This equation sums the USD values \( v_{\mathcal{x}, t_2} \) of each token \( \mathcal{x} \) within protocol \( p \) at time \( t_2 \), where \(\mathcal{x}\) is decided by the tokens that are present in the protocol at both the beginning and the end of the interval \( \tau \). 

To introduce edge weight, we define the value flow of a single token from \(p\) to \(q\) as:

\begin{equation}
    \label{eq:flow_pq}
    F_{pq,\tau}^{\mathcal{x}} = 
    \begin{cases} 
    \text{max}(0, -\Delta S_{p,\tau}^{\mathcal{x}}) & \text{if } \Delta S_{p,\tau}^{\mathcal{x}} < 0 \\
    \text{max}(0, \Delta S_{q,\tau}^{\mathcal{x}}) & \text{if } \Delta S_{q,\tau}^{\mathcal{x}} \ge 0 \\
    \end{cases}.
\end{equation}

This means the total value flow between any two protocols will remain positive. If negative, we reverse the flow direction. Based on this, we can define edge weight simply as: 

\begin{equation}
    w_{\tau}(e_{pq}) = \sum_{\mathcal{x} \in \mathcal{X}_{pq,\tau}} F_{\mathcal{x}}.
\end{equation}

Where \(q=\mathcal{M}\left(\mathcal{x}_{p, \tau}\right)\). This means that weight is calculated by considering all the token flows between \(p\) and \(q\), as indicated by \(\mathcal{M}\). Conceptually, the edge weight \(w_{\tau}(e_{pq})\) represents the change in credit exposure from protocol \(p\) to the issuing protocol \(q\) over the time interval \(\tau\). A positive edge weight indicates increasing exposure (tokens flowing from \(p\) to \(q\)), while the magnitude quantifies the extent of the exposure change. 

For single-token protocols, where each protocol is associated via \(\mathcal{M}\) with only one specific token \(\mathcal{x}\), the edge weights are directly equivalent to the flow of that token. Therefore, for such protocols, the flow \(F_{\mathcal{x}}\) for any edge \(e\) is equal to the weight of the edge \(w(e)\). This can be expressed as:

\begin{equation}
F_{\mathcal{x}}(e) = w(e).
\end{equation}

The procedure to calculate edge weights is described in Algorithm \ref{alg:edge_weights}.

    \begin{algorithm}[h]
        \caption{Calculate Edge Weights for Single Network Snapshot}
        \label{alg:edge_weights}
        \begin{algorithmic}[1]
        \Require $\mathcal{R}$: Mapping of protocols, chains, and tokens
        \Ensure Network $G = (P, E)$
        
        \Procedure{GenerateNetwork}{$\mathcal{R}$}
        \State Initialize $P \leftarrow \emptyset$, $E \leftarrow \emptyset$
        \State Let $\mathcal{P}$ be the set of all unique protocols and chains derived from $\mathcal{R}$
        \For{each protocol $p \in \mathcal{P}$}
            \State Aggregate $\sum n_{t_{\mathcal{x}}}^{(p,c)}$ and $\sum v_{t_{\mathcal{x}}}^{(p,c)}$ for all tokens $\mathcal{x}$ and chains $c$
        \EndFor
        \For{each chain $c \in \mathcal{C}$}
            \State Aggregate $\sum n_{t_{\mathcal{x}}}^{(c,p)}$ and $\sum v_{t_{\mathcal{x}}}^{(c,p)}$ for all tokens $\mathcal{x}$ and protocols $p$
        \EndFor
        \State $P \leftarrow P \cup \{\text{aggregated values}\}$
        \For{each $p \in P$}
            \State Compute node weight: $w(p) \leftarrow \sum_{\mathcal{x} \in \mathcal{X}_p} v_{t_{\mathcal{x}}}$
            \If{$w(p) < \theta$}
                \State $w(p) \leftarrow 0$
            \EndIf
        \EndFor
        \State $E \leftarrow$ \Call{PrepareEdge}{$\mathcal{R}$}
        \For{each link $(p, q) \in E$}
            \State Compute edge weight: $w(e_{pq}) \leftarrow \sum_{\mathcal{x} \in \mathcal{X}_{pq}} F_{\mathcal{x}}$
        \EndFor
        \State Format $P$ to include node attributes such as category
        \State \Return $G = (P, E)$
        \EndProcedure
        
        \Procedure{PrepareEdge}{$\mathcal{R}$}
        \State Initialize $\mathcal{E} \leftarrow \emptyset$
        \For{each pair $(p, c) \in \mathcal{R}$}
            \If{$\Delta S_{p, \tau}^\mathcal{x} < 0$}
                \State $\mathcal{E} \leftarrow \mathcal{E} \cup \{(p, q, F)\}$
            \Else
                \State $\mathcal{E} \leftarrow \mathcal{E} \cup \{(q, p, -F)\}$
            \EndIf
        \EndFor
        \State \Return $\mathcal{E}$
        \EndProcedure
        
        \end{algorithmic}
        \end{algorithm}

\FloatBarrier

\FloatBarrier
\section{Dataset Description}

In Table~\ref{tab:data_fields}, the DeXposure dataset provides daily snapshots of DeFi credit exposure networks in a structured JSON format. Each entry is indexed by date (e.g., \texttt{"2020-03-23"}) and contains two core components: a \texttt{nodes} list and a \texttt{links} list. The \texttt{nodes} field describes DeFi protocols or tokens, each represented by an identifier (\texttt{id}), the protocol’s total asset value in USD (\texttt{size}), and a detailed breakdown of token-level holdings (\texttt{composition}). The \texttt{links} field captures directed credit exposure relationships between protocols, specifying the source, target, exposure size in USD, and a token-level \texttt{composition} indicating the contribution of each asset to the exposure.

This dataset contains both balance-sheet information (via nodes) and inter-protocol dependencies (via links), enabling the reconstruction of market dynamic, multi-token credit exposure networks, and the temporal analysis of changes in the network. As of October 2025, the dataset covers more than 4,300 protocols, 602 blockchain networks, and 24,300 unique tokens, with all numerical values denominated in USD. JSON files are provided alongside CSV formats to support scalable downstream analysis.


\begin{table}[h!]
\centering
\caption{Dataset fields and descriptions for the DeXposure JSON network snapshots.}
\resizebox{\columnwidth}{!}{
\begin{tabular}{@{}llll@{}}
\toprule
Field & Type & Format / Unit & Description \\ \midrule

\texttt{date} & string & YYYY-MM-DD & Observation date (daily snapshot) \\

\texttt{nodes} & list & JSON array & List of protocol or token nodes in the network \\

\quad \texttt{id} & string & categorical & Unique identifier for the protocol or token \\

\quad \texttt{size} & float64 & USD & Total asset value held by the protocol \\

\quad \texttt{composition} & dict & \{token: amount\} & Token-level breakdown of holdings in USD \\

\texttt{links} & list & JSON array & Directed credit exposure relationships between nodes \\

\quad \texttt{source} & string & categorical & Origin protocol/token of the exposure \\

\quad \texttt{target} & string & categorical & Destination protocol/token of the exposure \\

\quad \texttt{size} & float64 & USD & Exposure value in USD \\

\quad \texttt{composition} & dict & \{token: amount\} & Token-level composition of the exposure \\

\texttt{network\_snapshot} & dict & JSON object & Full daily network containing nodes and links \\

\texttt{data\_source} & string & text & Data origin (e.g., on-chain, API, indexer) \\

\texttt{last\_updated} & datetime64[ns] & UTC ISO 8601 & Timestamp of dataset update \\

\bottomrule
\end{tabular}
}
\label{tab:data_fields}
\end{table}

\FloatBarrier
\section{Applications and Benchmarks}
We characterize the DeXposure dataset and demonstrate its utility through three benchmarks: (1) global network measurement capturing the structural evolution of credit exposure networks, (2) sector-level credit exposure dynamics during major market shocks (Terra and FTX), and (3) dynamic link prediction on temporal graph neural networks. These benchmarks illustrate how our dataset enables macro-level tracking and analysis of credit exposure evolution in the DeFi ecosystem.

\subsection{Graph Clustering for Global Network Measurement}
\label{sec:global_network_properties}

This benchmark characterizes the structural properties of the credit exposure network through edge composition analysis, network metrics, and visualization techniques.

\subsubsection{Problem Formulation}

Our objective is to characterize the structural evolution of the credit exposure network constructed in Section \ref{sec:network_model_description}. Recall that our dataset consists of a temporal sequence of weighted directed graphs \(\{G_\tau\}_{\tau \in T}\), where each snapshot \(G_\tau = (P_\tau, E_\tau)\) represents the credit exposure relationships between DeFi protocols at discrete time intervals. Each node \(p \in P_\tau\) represents a protocol with weight \(w(p)\) indicating its total value locked, and each directed edge \(e_{pq} \in E_\tau\) captures credit exposure from protocol \(p\) to protocol \(q\) with weight \(w(e_{pq})\) quantifying the exposure magnitude.

We aim to answer the following questions: (1) How does the network's structural complexity evolve over time? (2) What are the patterns of connectivity and centralization as the DeFi ecosystem matures? (3) Can we identify distinct clusters or communities of protocols, and how do these clusters change? To address these questions, we employ three complementary approaches: edge composition analysis to understand the token diversity underlying exposure relationships, network-level metrics to quantify structural properties such as centralization and clustering, and dimensionality reduction combined with clustering algorithms to visualize and identify protocol groupings in the high-dimensional feature space.

We calculate several network metrics to characterize the global properties of our credit exposure network and track its evolution over time:
\begin{enumerate}
    \item Degree Centralization:
    \begin{equation}
        DC = \frac{\sum_{i=1}^{|P_\tau|} (d_{max} - d_i)}{(|P_\tau| - 1)(|P_\tau| - 2)},
    \end{equation}
    where $d_{max}$ is the maximum total degree and $d_i$ is the total degree of protocol $i$. For our directed graph, we define the total degree as the sum of the in-degree and out-degree for each node.

    \item Degree Coefficient of Variation:
    \begin{equation}
        CV = \frac{\sigma_d}{\mu_d},
    \end{equation}
    where $\sigma_d$ and $\mu_d$ are the standard deviation and mean of the total degree distribution, respectively.

    \item Degree Distribution Entropy:
    \begin{equation}
        H = -\sum_{k} p(k) \log p(k),
    \end{equation}
    where $p(k)$ is the probability of a protocol having total degree $k$.

    \item Top 10\% Degree Concentration:
    \begin{equation}
        T = \frac{\sum_{i=1}^{0.1|P_\tau|} d_i}{\sum_{i=1}^{|P_\tau|} d_i},
    \end{equation}
    where total degrees are sorted in descending order.

    \item Assortativity:
    \begin{equation}
        r = \frac{\sum_{jk} jk(e_{jk} - q_j q_k)}{\sigma_q^2},
    \end{equation}
    where $e_{jk}$ is the joint probability distribution of the total degrees of two protocols at either end of a randomly chosen edge, and $q_k$ is the distribution of the total degree.

    \item Average Closeness Centrality:
    \begin{equation}
        CC = \frac{1}{|P_\tau|} \sum_{i=1}^{|P_\tau|} \frac{|P_\tau| - 1}{\sum_{j \neq i} d(i,j)},
    \end{equation}
    where $d(i,j)$ is the shortest path length between protocols $i$ and $j$. For our directed graph, we consider the shortest directed path from $i$ to $j$.
\end{enumerate}

\subsubsection{Edge Composition}

\begin{figure}[htbp]
    \centering
    \includegraphics[width=\columnwidth]{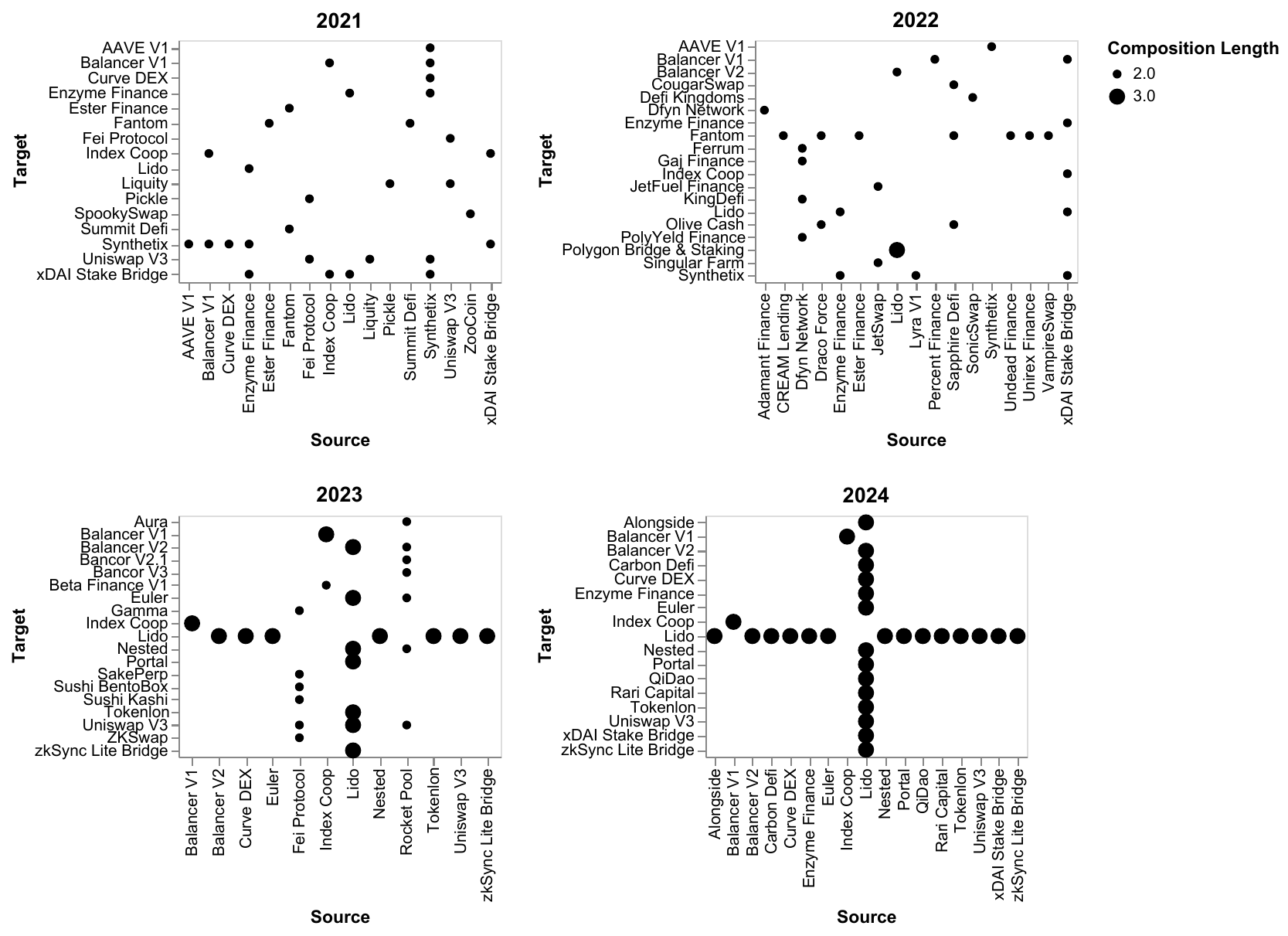}
    \caption{Select Edges with Highest Token Composition Length, Earliest Snapshot in the Year.}
    \label{fig:edge_composition}
\end{figure}

As edges between any two nodes are aggregated, there is no edge multiplicity within any snapshot. We define \textit{composition length} of an edge in the network as the number of distinct tokens that are associated with that edge. Therefore, for an edge \( e \) connecting protocols \( p \) and \( q \), the composition length is given by the cardinality of the set of tokens \( \mathcal{X}_{pq} \) associated with that edge, i.e., \( |\mathcal{X}_{pq}| \). Figure \ref{fig:edge_composition} shows the 30 edges with the highest token composition length in the earliest snapshot of each year, excluding the value transfers that are sourced to \textit{Unknown}.

\subsubsection{Network Analysis Results}

Table \ref{tab:network_analysis} provides an overview of the credit exposure network dynamics over four years from 2020 to 2024, indicating substantial expansion. The degree centralization peaked in 2021 before decreasing, suggesting an initial concentration followed by a more decentralized structure. The degree coefficient of variation increased from 2020 to 2021, then stabilized, indicating persistent inequality in connections. \added[id=H,comment=comment2]{The degree distribution entropy generally increased, suggesting growing complexity in connection patterns. The top 10\% degree concentration remained high after 2021, indicating that a small fraction of nodes consistently held a large share of connections. Assortativity remained negative but improved slightly, indicating diverse, interdependent connections. The average closeness centrality increased, suggesting improved network efficiency for exposure propagation.}

\begin{table}
\centering
\footnotesize
\caption{Summary of Network Metrics. This shows the earliest snapshot in the year, with weekly time granularity.}
\label{tab:network_analysis}
\begin{tabularx}{\columnwidth}{>{\centering\arraybackslash}p{0.5cm}XXXX}
\toprule
\multirow{2}{*}{Year} & \multicolumn{2}{c}{Network Size} & \multicolumn{2}{c}{Centralization} \\
\cmidrule(lr){2-3} \cmidrule(lr){4-5}
    & Nodes & Edges & Centrality & Variation* \\
\midrule
2020 & 9 & 8 & 0.389 & 1.237 \\
2021 & 388 & 516 & 0.597 & 6.029 \\
2022 & 3,655 & 12,835 & 0.736 & 4.888 \\
2023 & 5,872 & 27,220 & 0.737 & 4.477 \\
2024 & 8,539 & 45,817 & 0.744 & 4.759 \\
2025 & 11,087 & 69,710 & 0.744 & 4.918 \\
\midrule
\multirow{2}{*}{Year} & \multicolumn{2}{c}{Distribution} & \multicolumn{2}{c}{Connectivity} \\
\cmidrule(lr){2-3} \cmidrule(lr){4-5}
    & Entropy & Top 10\% & Assortativity & Closeness† \\
\midrule
2020 & 0.349 & 0.500 & -1.000 & 0.585 \\
2021 & 0.816 & 0.629 & -0.482 & 0.412 \\
2022 & 2.192 & 0.673 & -0.330 & 0.327 \\
2023 & 2.585 & 0.652 & -0.241 & 0.321 \\
2024 & 2.743 & 0.658 & -0.218 & 0.321 \\
2025 & 2.952 & 0.655 & -0.220 & 0.324 \\
\bottomrule
\end{tabularx}
*Degree Coefficient of Variation. †Average Closeness Centrality.
\end{table}

\begin{figure}[htbp]
    \centering
    \includegraphics[width=\columnwidth]{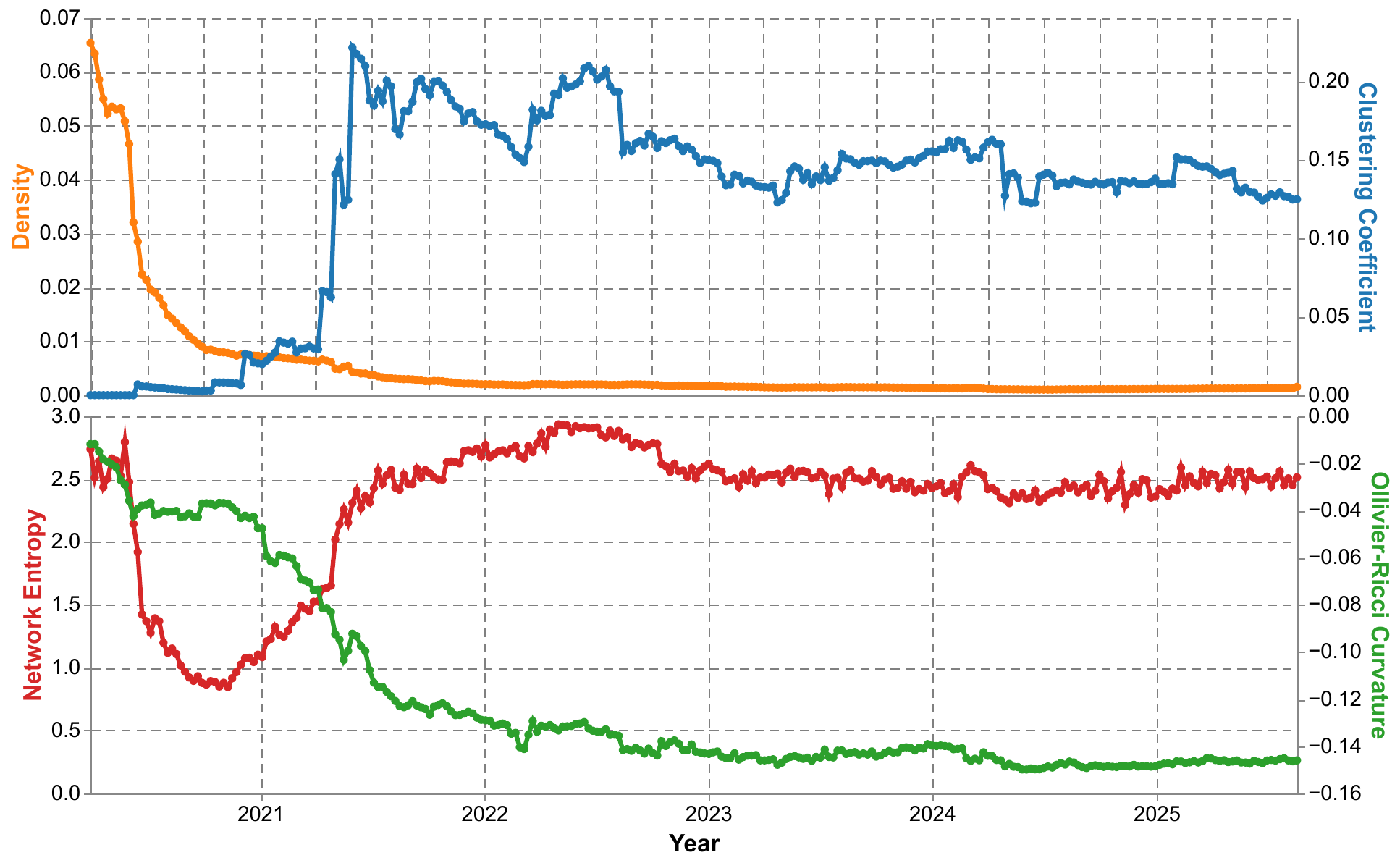}
    \caption{Network Density, Clustering Coefficient, Ollivier-Ricci Curvature, and Network Entropy Over Time}
    \label{fig:network_metrics_combined}
\end{figure}

In Figure \ref{fig:network_metrics_combined}, we observe trends in network density, clustering coefficients, network entropy, and Ollivier-Ricci curvature over time. The density decreases and stabilizes at a lower level, indicating a reduction in connections relative to possible maximums as the network expands. The clustering coefficient remains high, suggesting the preservation of local clustering properties and subgroup cohesion, a result of our stock-and-flow mapping. Network entropy stays within a narrow range due to consistent structural complexity. Ollivier-Ricci curvature fluctuates in negative values, indicating fluctuations in global connectivity. These metrics show our network snapshots decrease in density but maintain local clustering and complexity.

\subsubsection{Network Visualization and Clustering}

\added[id=B]{To understand the structural dynamics of the credit exposure network over time, we employ dimensionality reduction and clustering techniques. This allows us to visualize high-dimensional network data in a two-dimensional space and reveal protocol relationships and market structure that are not immediately apparent from raw network metrics.}

\begin{figure}[htbp]
    \centering
    \includegraphics[width=\columnwidth]{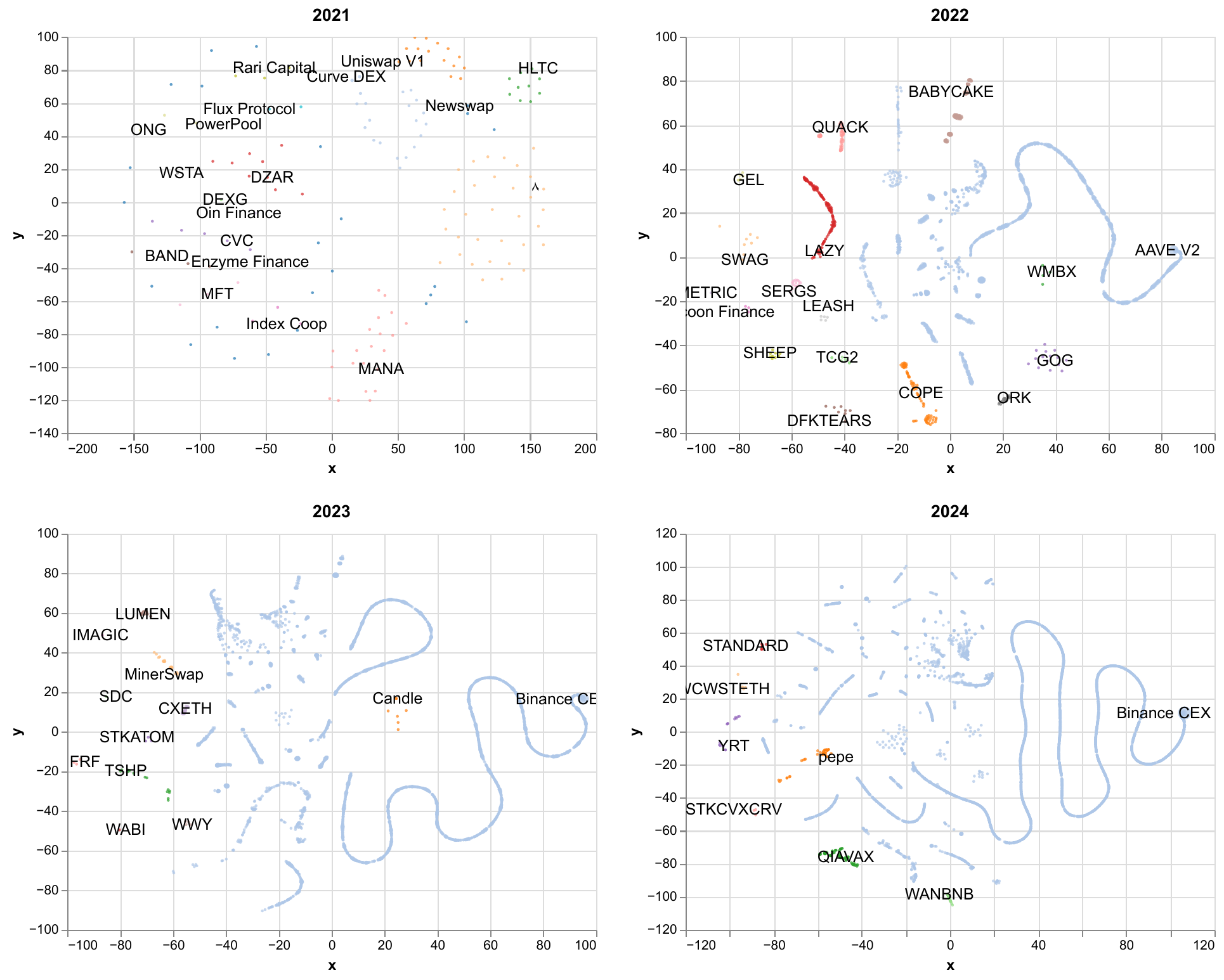}
    \caption{t-SNE Visualization of DeFi Protocols with DBSCAN Clustering (2021-2024)}
    \label{fig:tsne_clustering}
\end{figure}

\added[id=B]{Figure \ref{fig:tsne_clustering} illustrates the temporal evolution of the DeFi network structure from 2021 to 2024 using t-Distributed Stochastic Neighbor Embedding (t-SNE) \citep{Maaten2008-ul} visualization combined with Density-Based Spatial Clustering of Applications with Noise (DBSCAN) \citep{Ester1996-ou}. Each subplot represents a snapshot of the network for a specific year. The protocols are positioned in a two-dimensional space based on their network attributes, including degree centrality, betweenness centrality, eigenvector centrality, PageRank, clustering coefficient, and TVL. The size of each point corresponds to the protocol's TVL, while colors denote different clusters identified by DBSCAN.}

\added[id=B]{For the t-SNE visualization, we used a perplexity value of 30 and two dimensions for the output space to balance the preservation of both local and global structure in the high-dimensional data. The DBSCAN clustering algorithm was applied to the t-SNE output with parameters optimized for each year's data. For each year, we explored a range of epsilon values from 0.1 to 5.0 with a step of 0.1 and a minimum of 3 to 30 samples to identify clusters. The optimal parameters were selected based on the silhouette score, targeting between 5 and 20 clusters to capture a meaningful level of structure without over-segmentation.}

\added[id=B]{Analysis of these visualizations reveals the number of identified clusters decreases from 18 in 2021 to 8 in 2024, suggesting a consolidation of the DeFi ecosystem. Silhouette scores, which measure how similar an object is to its own cluster compared to other clusters, decline from 0.52 to -0.02, indicating increasing overlap between clusters. This pattern reveals an evolution from distinct protocol clusters towards a more integrated and complex structure where protocols increasingly share features across boundaries. The figure also highlights the dominant protocols within each cluster, with the largest share represented by market leaders.}

\added[id=B,comment=comment5]{This measurement reveals three key trends: (1) market structure consolidation, evidenced by fewer clusters; (2) increasing ecosystem integration and complexity, shown by decreasing silhouette scores and greater cluster overlap; and (3) emergence of dominant protocols within specific market segments. These indicate a maturing DeFi market with increased interconnectedness between protocol types and established market leaders, while maintaining overall structural diversity.}

\subsection{Vector Autoregression for Market Shocks Analysis} 

\label{sec:value_flows}

\subsubsection{Problem Formulation}

We examine and compare two market events that had substantial impacts on the DeFi ecosystem: the collapses of Terra (May 9, 2022) and FTX (November 7, 2022). Our objective is to characterize how these shocks propagate through the credit exposure network at both sector and protocol levels, and to quantify the dynamic interactions between sectoral exposure shifts.

We begin by aggregating the network into sectors based on protocol functionality categories from DeFiLlama \cite{DefiLlama2021-je}. We focus on five key sectors: Infrastructure (protocols providing basic blockchain services and cross-chain bridges), Services \& Financial Products (insurance, derivatives, and financial tools), Lending, Borrowing \& Real World Assets (protocols facilitating lending and borrowing of digital and tokenized real-world assets), \added[id=B,comment=comment10]{Asset Management} (protocols managing investment portfolios and yield farming strategies), and Trading \& Exchanges (decentralized exchanges and automated market makers facilitating token swaps).

For each sector, we define an exposure shift ratio to quantify the direction and magnitude of exposure changes:
\begin{equation}
\rho=\frac{F_i - F_o}{F_i + F_o},
\end{equation}
where \(F_i\) and \(F_o\) represent exposure expansion and contraction values. Values range from -1 to 1, with -1 indicating exclusive contraction (negative market conditions), 0 indicating balanced shifts (neutral conditions), and 1 indicating exclusive expansion (positive conditions)\footnote{This approach draws on fund flow analysis in traditional finance \cite{Frazzini2005-iw} and network finance studies linking flow patterns to market conditions and systemic risk \cite{Squartini2013-qu, Codd2017-qq}.}.

To quantify the dynamic interactions between sector expansion and contraction over time, we employ a Vector Autoregression (VAR) model. VAR captures the interdependencies between multiple time series variables, allowing us to analyze how shocks to one sector's exposure propagate to others. We present Impulse Response Functions (IRF) derived from this model, which illustrate how an exposure expansion to a sector affects other variables over time, revealing potential contagion pathways across the DeFi ecosystem.

\subsubsection{Cross-Sectoral Exposure Dynamics}

\begin{figure}[htbp]
    \centering
    \includegraphics[width=\columnwidth]{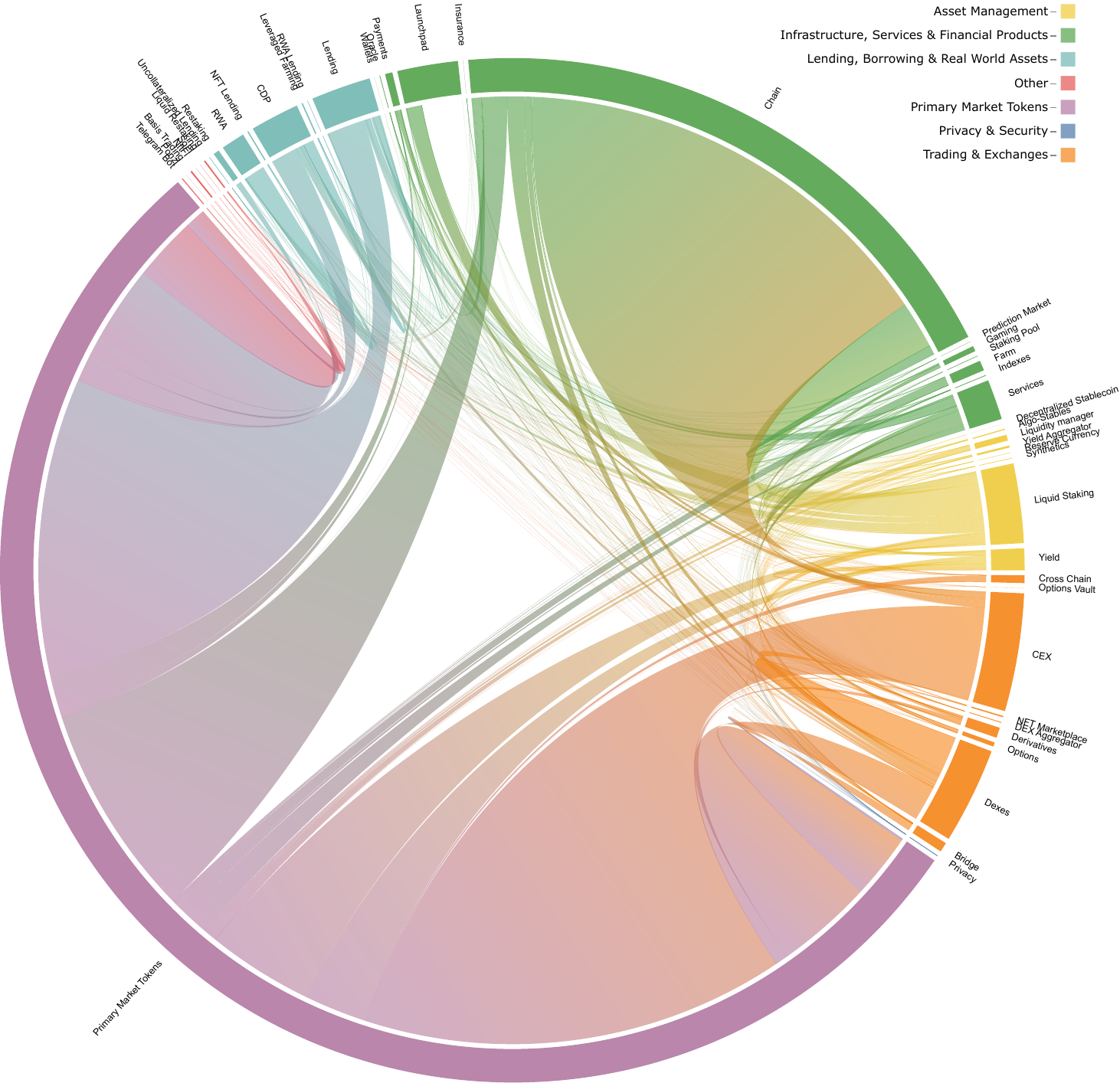}
    \caption{Chord Diagram of Exposure Shifts Across Sectors. Network data taken from time snapshot 2024-06-01. Nodes and edges are aggregated by categories. Sectors indicated by colors. Magnitude indicated by the USD value of cross-category exposure shifts.}
    \label{fig:combined_legend_for_cross_sector_flow}
\end{figure}

Figure \ref{fig:combined_legend_for_cross_sector_flow} provides an overview of cross-sectoral exposure dynamics by aggregating network nodes and edges based on sector categories.

\FloatBarrier

\begin{figure}[htbp]
    \centering
    \includegraphics[width=\columnwidth]{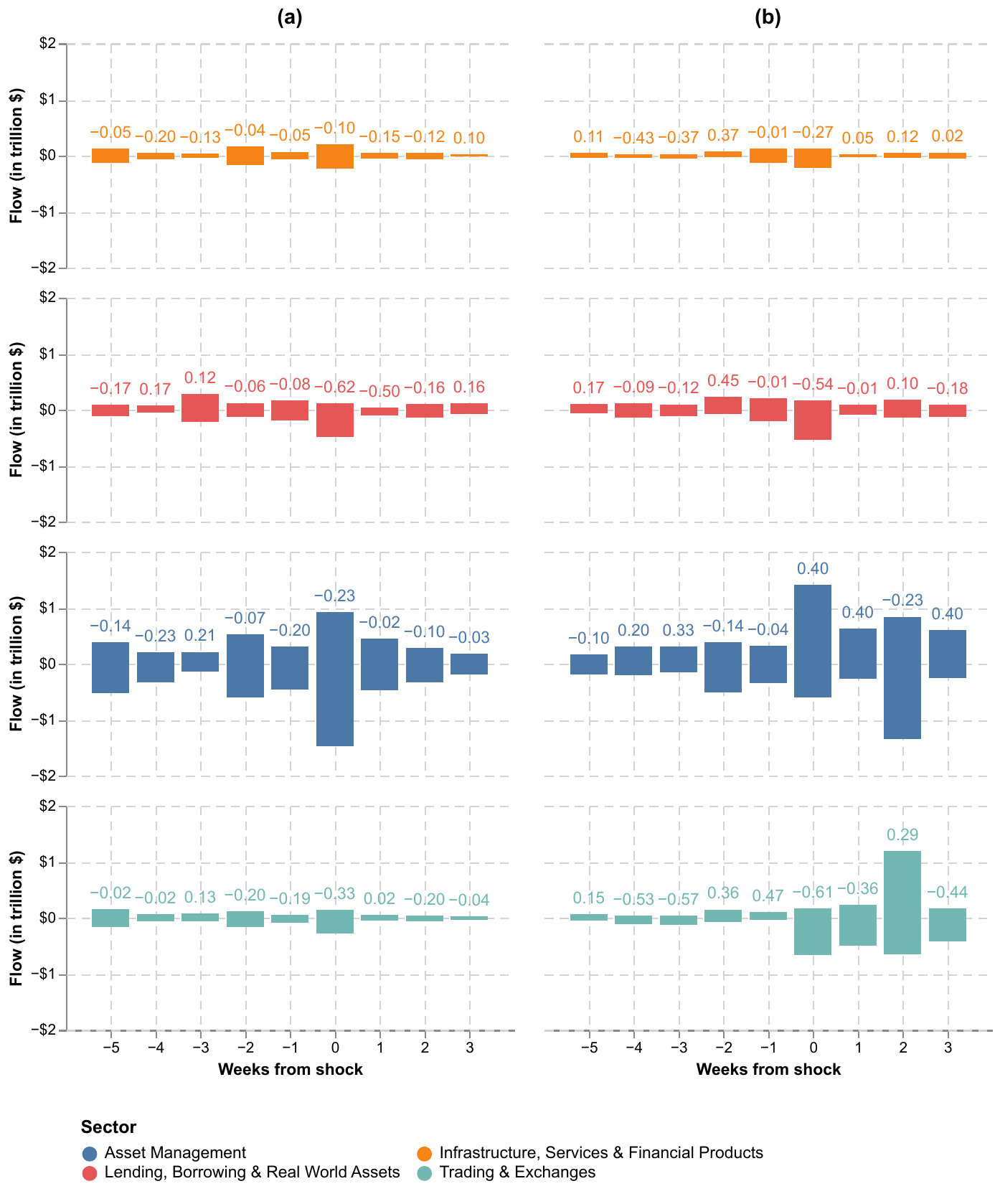}
    \caption{Sectoral Exposure Shifts During Major Market Shocks. (a) Terra (May 9, 2022). (b) FTX (November 7, 2022). Values show the changes of \(\rho\) over time.}
    \label{fig:combined_incidents_altair}
\end{figure}

Figure \ref{fig:combined_incidents_altair} illustrates the impacts of the Terra and FTX incidents on the given DeFi sectors. The Terra incident caused more widespread disruption across the ecosystem, with two sectors, 1. Asset Management, 2. Lending, Borrowing \& Real World Assets experienced the largest weekly exposure contraction of approximately 1.47 trillion and 0.49 trillion dollars during the incident week. In contrast, the FTX incident's impact was more concentrated, mainly affecting the Trading \& Exchanges sector, which experienced exposure contraction of about 0.67 trillion dollars in the incident week. It should also be noted that the Asset Management sector recorded a net expansion, receiving more exposure than it lost.

The figure shows divergent recovery patterns: following the FTX incident, most sectors recovered within 2 weeks, as indicated by expansion exceeding contraction. The Terra incident, however, resulted in a longer recovery period, with some sectors taking up to 4 weeks to show similar recovery signals. These variations in impact and recovery likely stem from the fundamental differences between the two events: Terra's collapse transmitted to DeFi protocols and their interconnected ecosystem, while FTX's failure was confined to a centralized exchange. 

\begin{figure}[htbp]
\centering
\includegraphics[width=\columnwidth]{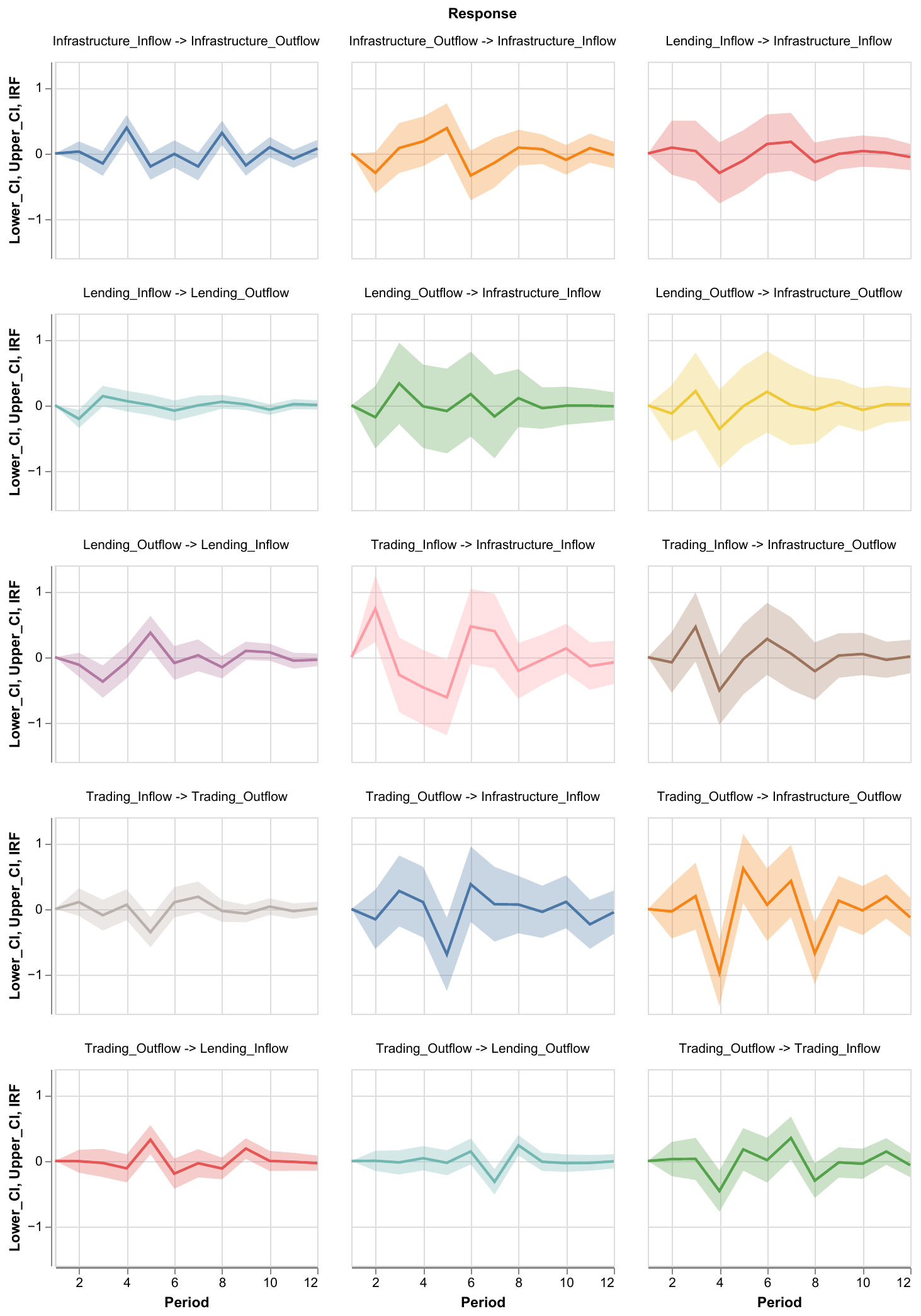}
\caption{IRF of Select Sectors. Showing expansion and contraction over 12 periods, where each period is week.}
\label{fig:irf_plots_with_independent_y_axes}
\end{figure}

Figure \ref{fig:irf_plots_with_independent_y_axes} presents the Impulse Response Functions (IRF) derived from our VAR model. \added[id=B,comment=comment9]{The IRF analysis reveals potential contagion pathways across the DeFi ecosystem. For instance, a positive shock to Trading \& Exchanges exposure expansion may lead to increased exposure contraction from Lending platforms, indicating a shift in capital allocation preferences during market events.}

\subsubsection{Protocol-Level Network Dynamics}

To complement our sector-level analysis, we examine protocol-specific behaviors during major market events. Table \ref{tab:incident_analysis} presents weekly network dynamics during the Terra and FTX shocks. This protocol-level perspective reveals irregular patterns that may indicate significant market movements or potential risks. For instance, in the fourth week following the Terra incident, we observe immense exposure shifts from the Reverse protocol to the Defi ecosystem. In Section \ref{sec:dynamic_link_prediction}, we discuss methods to mine and predict such insights at scale systematically.

We quantify protocol influence by calculating the net exposure change for each node in the network, defined as the difference between total exposure expansion and contraction. Critical network connections are identified through cut edges, whose removal would increase the number of connected components in the graph. For the Asset Management and Trading \& Exchanges sectors, we report the top protocol by absolute net exposure change and the cut edge with the most significant transaction volume. 

\begin{table}[htbp]
    \centering
    \footnotesize
    \caption{Protocol-level analysis during Terra and FTX. For each week, the first row represents the Asset Management sector, and the second row represents the Trading \& Exchanges sector. We report the top protocol by absolute net exposure change and the largest cut edge by transaction volume. Values are in billions of dollars.}
    \label{tab:incident_analysis}
    \begin{tabularx}{\columnwidth}{>{\centering\arraybackslash}p{0.4cm}l>{\raggedright\arraybackslash}p{1cm}X>{\raggedright\arraybackslash}p{0.6cm}}
    \toprule
    Week & Protocol & Value & Cut Edge & Value \\
    \midrule
    \multirow{2}{*}{-4} & Lido & -1.73 & Yearn $\rightarrow$ USDT & 0.19 \\
    & Stargate V1 & -0.83 & UniswapV2 $\rightarrow$ USDT & 0.10 \\
    \cmidrule{1-5}
    \multirow{2}{*}{-3} & Alchemix & -0.42 & Timewarp $\rightarrow$ Volta Club & 0.13 \\
    & Curve & 2.30 & crvUSD $\rightarrow$ Curve & 2.30 \\
    \cmidrule{1-5}
    \multirow{2}{*}{-2} & Lido & -1.27 & Yearn $\rightarrow$ USDT & 0.18 \\
    & Curve & -2.61 & Curve $\rightarrow$ crvUSD & 2.61 \\
    \cmidrule{1-5}
    \multirow{2}{*}{-1} & Lido & -2.32 & Yearn $\rightarrow$ USDT & 0.18 \\
    & Curve & 0.69 & RenVM $\rightarrow$ renBTC & 0.34 \\
    \cmidrule{1-5}
    \multirow{2}{*}{0} & Convex & -5.54 & Yearn $\rightarrow$ USDT & 0.66 \\
    & Curve & -3.50 & Curve $\rightarrow$ crvUSD & 3.50 \\
    \cmidrule{1-5}
    \multirow{2}{*}{1} & Lido & 0.91 & Yearn $\rightarrow$ USDT & 0.07 \\
    & Curve & -0.99 & WBTC $\rightarrow$ WBTC & 0.27 \\
    \cmidrule{1-5}
    \multirow{2}{*}{2} & Benqi & -0.41 & Timewarp $\rightarrow$ Volta Club & 0.14 \\
    & WBTC & -0.41 & WBTC $\rightarrow$ WBTC & 0.41 \\
    \cmidrule{1-5}
    \multirow{2}{*}{3} & Lido & -0.51 & Yearn $\rightarrow$ USDT & 0.02 \\
    & Curve & -0.26 & Tinyman $\rightarrow$ USDT & 0.04 \\
    \cmidrule{1-5}
    \multirow{2}{*}{4} & Reverse & -5.71 & Yearn $\rightarrow$ USDT & 0.10 \\
    & Defi Kingdoms & 5.71 & Curve $\rightarrow$ crvUSD & 1.29 \\
    \midrule
    \multirow{2}{*}{-4} & Arrakis V1 & -0.42 & Marinade $\rightarrow$ Solana & 0.04 \\
    & Curve & -1.04 & Thor* $\rightarrow$ Bepswap & 0.03 \\
    \cmidrule{1-5}
    \multirow{2}{*}{-3} & Lido & 0.21 & Liquity $\rightarrow$ LUSD & 0.05 \\
    & Uniswap V3 & -0.86 & Ultron $\rightarrow$ UltronSwap & 0.05 \\
    \cmidrule{1-5}
    \multirow{2}{*}{-2} & Rocket Pool & 0.68 & Convex $\rightarrow$ MUSD3CRV & 0.01 \\
    & Curve & 0.98 & WBTC $\rightarrow$ WBTC & 0.30 \\
    \cmidrule{1-5}
    \multirow{2}{*}{-1} & Rocket Pool & 0.16 & KLAY $\rightarrow$ Stake.ly & 0.05 \\
    & Curve & 0.58 & WBTC $\rightarrow$ WBTC & 0.20 \\
    \cmidrule{1-5}
    \multirow{2}{*}{0} & Convex & -1.81 & Volta Club $\rightarrow$ Timewarp & 0.05 \\
    & Curve & -10.00 & SUNSwap V1 $\rightarrow$ USDT & 0.27 \\
    \cmidrule{1-5}
    \multirow{2}{*}{1} & Lido & -0.40 & Liquity $\rightarrow$ LUSD & 0.07 \\
    & Binance & 5.24 & Bepswap $\rightarrow$ Thor* & 0.02 \\
    \cmidrule{1-5}
    \multirow{2}{*}{2} & Arrakis V1 & -0.25 & Liquity $\rightarrow$ LUSD & 0.06 \\
    & Binance & 9.75 & WBTC $\rightarrow$ WBTC & 0.20 \\
    \cmidrule{1-5}
    \multirow{2}{*}{3} & Lido & -0.23 & WETH $\rightarrow$ Frax Ether & 0.01 \\
    & Binance & -0.87 & WBNB $\rightarrow$ BNB & 0.05 \\
    \cmidrule{1-5}
    \multirow{2}{*}{4} & ApeCoin & 0.14 & Liquity $\rightarrow$ LUSD & 0.03 \\
    & Binance & -5.37 & UniswapV2 $\rightarrow$ USDT & 0.02 \\
    \bottomrule
    \end{tabularx}
    \smallskip
    \raggedright
    \footnotesize
    *Thor: Thorchain. Benqi: Benqi Staked Avax. Convex: Convex Finance. Binance: Binance CEX.
\end{table}

\subsection{Temporal Graph Neural Networks for Dynamic Link Prediction}
\label{sec:dynamic_link_prediction}

\subsubsection{Problem Formulation}

Dynamic link prediction is a common task in the analysis of temporal networks. It is pertinent in the DeFi context, where the network's topology can change rapidly due to market dynamics or other external factors. By forecasting the formation or dissolution of credit exposure links based on historical network data, we can gain insights into the evolving network structure and anticipate future changes. 

Here we formalize the dynamic link prediction problem for our temporal graphs, each denoted as \( G_\tau \). Let \( \mathcal{G} = \{G_\tau\}_{\tau=t_1-t_0}^{|T|} \) be a sequence of graph snapshots. Given the discrete nature of our networks, the objective of dynamic link prediction is to predict the edge set \( E_{\tau+1} \) of the graph \( G_{\tau+1} \) given the previous graph snapshots \( \{G_{\tau_1}, G_{\tau_2}, \ldots, G_{\tau_{|T|}}\} \). This involves predicting both the formation of new edges and the dissolution of existing ones.

\subsubsection{Model Description}

We employ a temporal graph neural network (TGNN) model based on the ROLAND framework \cite{You2022-oy,Dileo2024-ep} that leverages node features and edge dynamics to predict future connections. We give a high-level overview of our implementation along three components: node representation learning, edge prediction, and temporal dynamics.

\begin{enumerate}
    
    \item At each time interval \( \tau \), node features as the input space \( \mathbf{x}_\tau^p = [w(p)]\) for each node \( p \in P_\tau \) are initially processed using Multi-Layer Perceptron (MLP) to enhance the feature vector, which is then updated based on the previous features and the structural information from the graph \( G_\tau \) using a Graph Convolutional Network (GCN) layer:
    \begin{equation}
    \mathbf{h}_\tau^p = \text{GCN}(\text{MLP}(\mathbf{x}_\tau^p), \{ \mathbf{h}_{\tau-1}^u : u \in \mathcal{N}(p) \}),
    \end{equation}
    where \( \mathcal{N}(p) \) denotes the neighbors of node \( p \) in graph \( G_\tau \), and \( \mathbf{h}_{\tau-1}^u \) are the representations of these neighbors from the previous time interval.

    \item For each potential edge \( (p, q) \) at time \( \tau+1 \), the model computes a score \( s_{\tau+1}^{p,q} \) indicating the likelihood of an edge forming or persisting between \( p \) and \( q \). This score is calculated using the dot product of the node representations, followed by a sigmoid activation:
    \begin{equation}
    s_{\tau+1}^{p,q} = \sigma(\mathbf{h}_\tau^p \cdot \mathbf{h}_\tau^q),
    \end{equation}
    where \( \sigma \) is the sigmoid activation function, and \( \cdot \) denotes the dot product.

    \item We incorporate temporal dynamics by using a Gated Recurrent Unit (GRU) to update the node representations over time, which allows the model to maintain a memory of past node states and predict future connections:
    \begin{equation}
    \mathbf{h}_\tau^p = \text{GRU}(\text{GCN}(\mathbf{h}_\tau^p), \mathbf{h}_{\tau-1}^p),
    \end{equation}
    where \( \mathbf{h}_{\tau-1}^p \) are the representations of node \( p \) from the previous time interval, and \( \text{GCN}(\mathbf{h}_\tau^p) \) represents the current processed state of node \( p \).
\end{enumerate}

For replicability, we provide an overview of our model setup in Table \ref{tab:nn_model_parameters}. 

\begin{table}[htbp]
\centering
\footnotesize
\caption{Configuration and Hyperparameters.}
\begin{tabularx}{\columnwidth}{X>{\hsize=2\hsize}X>{\hsize=0.5\hsize}X>{\hsize=0.5\hsize}X}
\toprule
Category & Parameter & \multicolumn{2}{c}{Value*} \\
\midrule
\multirow{3}{*}{Learning} & Epochs Per Snapshot & \multicolumn{2}{c}{50} \\
& Early-Stop Tolerance & \multicolumn{2}{c}{0.0001} \\
& Learning Rate & \multicolumn{2}{c}{0.01} \\
\midrule
\multirow{4}{*}{Architecture} & MLP1  & 1† & 256 \\
& MLP2  & 256 & 128 \\
& GCN1  & 128 & 64 \\
& GCN2  & 64 & 32 \\
\midrule
\multirow{2}{*}{Update} & GRU1  & 64 & 64 \\
& GRU2 & 32 & 32 \\
\bottomrule
\end{tabularx}
\smallskip
\raggedright
\footnotesize
*In-Out Channels. †Depth of Node Features.
\label{tab:nn_model_parameters}
\end{table}

\subsubsection{Training and Evaluation}

We train on our weekly historical dataset of graph snapshots by minimizing a cross-entropy loss function that measures the discrepancy between the predicted edge probabilities and the actual edge formations. 

The model's performance is evaluated using the Area Under the Precision-Recall Curve (AUPRC) metric. This metric is beneficial for datasets with imbalanced classes, which is a common scenario in link prediction tasks. In such tasks, there are fewer actual positive links than potential negative links; i.e., existing links are typically much fewer than non-existing links.

\begin{figure}[ht]
\centering
\includegraphics[width=\columnwidth]{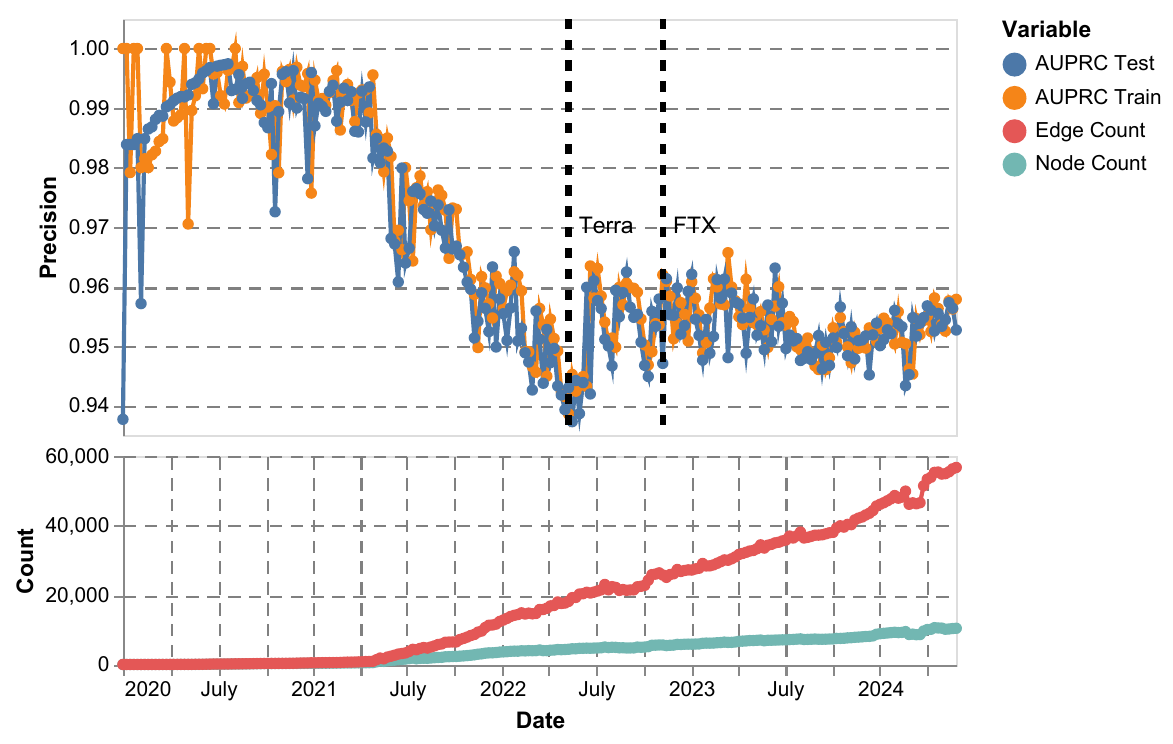}
\caption{Variation in AUPRC Precision and Network Topology. Across all network snapshots, the above plot shows prediction accuracy, and the plot below shows node and edge counts.}
\label{fig:network_analysis_over_time}
\end{figure}

In other words, the AUPRC metric is a measure of the model's ability to retrieve relevant instances across all possible recall levels. It is defined as the integral of the precision-recall curve, which can be approximated by the weighted mean of precisions achieved at each threshold:

\begin{equation}
\text{AUPRC} = \sum_{k=1}^{n} \text{Precision}(k) \Delta r(k),
\end{equation}

where \( \text{Precision}(k) \) is the precision at the \( k \)-th threshold, and \( \Delta r(k) \) is the change in recall from the \( (k-1) \)-th to the \( k \)-th threshold. The higher the AUPRC, the better the model is at handling the positive class in the presence of a large number of negative classes. We run our evaluation across all the snapshots and present our results in Figure \ref{fig:network_analysis_over_time}.

The observed decline in prediction accuracy correlates well with major market shocks. Note that this occurs independently of the consistent growth in network nodes and edges, which suggests that the AUPRC metric captures underlying network dynamics not solely attributable to changes in network size. This observation opens up potential avenues for further research into using network-based metrics as indicators of market conditions.

\subsection{Financial Insights Gained from DeXposure}

Our analysis of the DeXposure dataset reveals several key insights into the DeFi market structure and dynamics.

First, in Section \ref{sec:global_network_properties}, the network's centralization dynamics have shifted over time. After peaking in 2021, degree centralization decreased, suggesting an initial concentration followed by a more distributed structure. However, the persistent high concentration of top 10\% degrees indicates that a small set of nodes consistently holds a large share of connections, suggesting the presence of influential protocols within the ecosystem.

Second, also in Section \ref{sec:global_network_properties}, structural complexity has increased, as evidenced by rising degree distribution entropy and stable network entropy. Despite decreasing network density, the high clustering coefficient indicates strong local interconnections. The negative but improving assortativity suggests diverse, interdependent connections between protocols, reflecting the ecosystem's maturation. Visualization techniques reveal a consolidation of the DeFi ecosystem from 2021 to 2024, with increasing integration and complexity. The emergence of dominant protocols within specific market segments indicates a maturing market with established leaders, while maintaining overall structural diversity.

Third, in Section \ref{sec:value_flows}, our analysis of exposure shifts across sectors during major market events illustrates that the observed reversal in expansion and contraction dynamics post-incidents warrants further investigation and suggests adaptive responses within the ecosystem.

Fourth, in Section \ref{sec:dynamic_link_prediction}, the dynamic link prediction model's performance correlates with major market shocks. This opens up new possibilities for market condition indicators, which could be valuable tools for analysis and market participants by aiding in monitoring and predicting DeFi market dynamics.

Finally, in Section \ref{sec:network_model_description}, our stock and flow approach to measuring inter-protocol credit exposure dynamics through TVL data reconstructs market activity on a macro scale and provides a more comprehensive view of exposure changes within DeFi. By tracking changes in token holdings across protocols, we reveal the temporal evolution of credit exposure networks in DeFi. This enables analysis at a scale previously unseen with transaction-based approaches.

These findings can directly support DeFi market analysis and monitoring for stakeholders. By providing insights on DeFi exposure dynamics, market structure evolution, and systemic risk factors, our research offers a comprehensive and interpretable overview of the DeFi ecosystem. This enhanced understanding can improve the ability of these stakeholders to monitor and predict market dynamics.
\FloatBarrier
\section{Conclusion}

This paper presents the DeXposure dataset, a comprehensive resource for analyzing inter-protocol credit exposure in decentralized finance through a network-based approach using TVL data. By constructing temporal snapshots of credit exposure relationships, we enable large-scale analysis of DeFi ecosystem dynamics previously inaccessible through transaction-level data alone. Our benchmarks demonstrate the dataset's utility for tracking network evolution, analyzing market shocks, and predicting future connections through machine learning models.


We envision building on the DeXposure dataset to develop machine learning technologies for DeFi. For example, graph-based generative models can be used to simulate extreme market events, contagion dynamics, or liquidity shocks in DeFi ecosystems. Another promising direction is to fine-tune foundation models, especially emerging graph foundation models, on the DeXposure dataset to improve domain-specific tasks in DeFi, such as protocol interaction analysis, anomaly detection, link prediction, and systemic risk forecasting.

We publicly release our dataset and code at \url{https://github.com/dthinkr/DeXposure}. We also present our results through an interactive DeFi digital tool at \url{https://ccaf.io/defi/ecosystem-map/visualisation/graph}.
\FloatBarrier
\section*{Acknowledgements}

We express our gratitude to the DefiLlama team for providing the open-source data that made this research possible. We also thank the research team at the Cambridge Centre for Alternative Finance (CCAF) for their valuable feedback and support throughout the development of the DeXposure dataset. This work was supported by the Cambridge Digital Assets Programme.

\FloatBarrier



\vskip 0.2in
\bibliography{ref}

\begin{thebibliography}{56}
\providecommand{\natexlab}[1]{#1}
\providecommand{\url}[1]{\texttt{#1}}
\expandafter\ifx\csname urlstyle\endcsname\relax
  \providecommand{\doi}[1]{doi: #1}\else
  \providecommand{\doi}{doi: \begingroup \urlstyle{rm}\Url}\fi

\bibitem[Alamsyah and Muhammad(2024)]{Alamsyah2024-mx}
Andry Alamsyah and Ivan~Farid Muhammad.
\newblock {Unraveling the crypto market: A journey into decentralized finance
  transaction network}.
\newblock \emph{Digital Business}, 4\penalty0 (100074):\penalty0 100074,
  February 2024.
\newblock ISSN 2666-9544.
\newblock \doi{10.1016/j.digbus.2024.100074}.
\newblock URL \url{http://dx.doi.org/10.1016/j.digbus.2024.100074}.

\bibitem[Alhaidari et~al.(2025)Alhaidari, Kalal, Palanisamy, and
  Sural]{Alhaidari2025}
Abdulrahman Alhaidari, Bhavani Kalal, Balaji Palanisamy, and Shamik Sural.
\newblock {SolRPDS}: A dataset for analyzing rug pulls in solana decentralized
  finance.
\newblock In \emph{Proceedings of the Fifteenth ACM Conference on Data and
  Application Security and Privacy ({CODASPY} '25)}, pages 293--298, 2025.
\newblock \doi{10.1145/3714393.3726487}.

\bibitem[Aramonte et~al.(2021)Aramonte, Huang, and Schrimpf]{aramonte2021defi}
Sirio Aramonte, Wenqian Huang, and Andreas Schrimpf.
\newblock Defi risks and the decentralisation illusion.
\newblock \emph{BIS Quarterly Review}, pages 21--36, December 2021.
\newblock URL \url{https://www.bis.org/publ/qtrpdf/r_qt2112b.htm}.

\bibitem[Auer et~al.(2024)Auer, Haslhofer, Kitzler, Saggese, and
  Victor]{auer2024tech_defi}
Raphael Auer, Bernhard Haslhofer, Stefan Kitzler, Pietro Saggese, and Friedhelm
  Victor.
\newblock The technology of decentralized finance (defi).
\newblock \emph{Digital Finance}, 6\penalty0 (1):\penalty0 55--95, 2024.
\newblock \doi{10.1007/s42521-023-00088-8}.
\newblock URL \url{https://www.bis.org/publ/work1066.htm}.

\bibitem[Auer et~al.(2025)Auer, Borio, et~al.]{auer2025crypto_functions}
Raphael Auer, Claudio Borio, et~al.
\newblock Cryptocurrencies and decentralised finance: Functions and financial
  stability implications.
\newblock BIS Papers 156, Bank for International Settlements, 2025.
\newblock URL \url{https://www.bis.org/publ/bppdf/bispap156.htm}.

\bibitem[Badev and Watsky(2023)]{badev2023interconnected_defi}
Anton Badev and Ethan Watsky.
\newblock Interconnected defi: Ripple effects from the terra collapse.
\newblock Feds notes, Board of Governors of the Federal Reserve System, 2023.
\newblock URL
  \url{https://www.federalreserve.gov/econres/notes/feds-notes/interconnected-defi-ripple-effects-from-the-terra-collapse-20231115.html}.

\bibitem[{Basel Committee on Banking Supervision}(2023)]{BCBS-counterparty}
{Basel Committee on Banking Supervision}.
\newblock {CRE50} - counterparty credit risk definitions and terminology.
\newblock Technical report, Bank for International Settlements, 2023.
\newblock Basel Framework.

\bibitem[Buterin(2014)]{Buterin2014-ethereum}
Vitalik Buterin.
\newblock Ethereum: A next-generation smart contract and decentralized
  application platform.
\newblock Ethereum Whitepaper, 2014.
\newblock URL \url{https://ethereum.org/en/whitepaper/}.

\bibitem[{Cambridge Center for Alternative
  Finance}(2024)]{Cambridge-Center-for-Alternative-Finance2024-fi}
{Cambridge Center for Alternative Finance}.
\newblock {CCAF Digital Tools - Cambridge Centre for Alternative Finance}.
\newblock Technical report, Cambridge Center for Alternative Finance, 2024.
\newblock URL \url{https://ccaf.io/}.

\bibitem[Carpentier-Desjardins et~al.(2025)Carpentier-Desjardins,
  Paquet-Clouston, Kitzler, and Haslhofer]{Carpentier2025}
Catherine Carpentier-Desjardins, Masarah Paquet-Clouston, Stefan Kitzler, and
  Bernhard Haslhofer.
\newblock Mapping the {DeFi} crime landscape: An evidence-based picture.
\newblock \emph{Journal of Cybersecurity}, 11\penalty0 (1), 2025.
\newblock \doi{10.1093/cybsec/tyae029}.

\bibitem[Chan-Lau(2018)]{chanlau2018systemic_communities}
Jorge~A. Chan-Lau.
\newblock Systemic centrality and systemic communities in financial networks.
\newblock \emph{Journal of Risk Management in Financial Institutions},
  11\penalty0 (2):\penalty0 148--163, 2018.

\bibitem[Chemaya et~al.(2025)Chemaya, Cong, Jorgensen, Liu, and
  Zhang]{Chemaya2025}
Nir Chemaya, Lin~William Cong, Emma Jorgensen, Dingyue Liu, and Luyao Zhang.
\newblock A dataset of {Uniswap} daily transaction indices by network.
\newblock \emph{Scientific Data}, 12\penalty0 (1):\penalty0 93, 2025.
\newblock \doi{10.1038/s41597-024-04042-0}.

\bibitem[Chen and Tsai(2025)]{Chen2025}
Chi-Sheng Chen and Aidan Hung-Wen Tsai.
\newblock Benchmarking classical and quantum models for {DeFi} yield prediction
  on curve finance.
\newblock \emph{arXiv preprint}, 2025.

\bibitem[Codd(2017)]{Codd2017-qq}
Brinley Codd.
\newblock {The decline of solvency contagion risk}.
\newblock \emph{Bank of England Working Papers}, June 2017.
\newblock URL \url{https://ideas.repec.org//p/boe/boeewp/0662.html}.

\bibitem[Conlon et~al.(2024)Conlon, Corbet, and
  Sch{\"u}tze]{conlon2024ftx_contagion}
Thomas Conlon, Shaen Corbet, and Rainer Sch{\"u}tze.
\newblock Contagion effects of permissionless worthless cryptocurrency tokens:
  Evidence from the collapse of ftx.
\newblock \emph{Journal of International Financial Markets, Institutions and
  Money}, 87:\penalty0 101838, 2024.
\newblock \doi{10.1016/j.intfin.2023.101838}.

\bibitem[{DefiLlama}(2021{\natexlab{a}})]{DefiLlama2021-je}
{DefiLlama}.
\newblock {projects/helper/tokenMapping.js at main ·
  DefiLlama/DefiLlama-Adapters}, 2021{\natexlab{a}}.
\newblock URL
  \url{https://github.com/DefiLlama/DefiLlama-Adapters/blob/main/projects/helper/tokenMapping.js}.

\bibitem[{DefiLlama}(2021{\natexlab{b}})]{DefiLlama2021-pz}
{DefiLlama}.
\newblock {API Docs - {DefiLlama}}.
\newblock \url{https://defillama.com/docs/api}, 2021{\natexlab{b}}.
\newblock Accessed: 2024-6-6.

\bibitem[{DefiLlama}(2024)]{DefiLlama2024-zw}
{DefiLlama}.
\newblock {projects/ethena/index.js at main · DefiLlama/DefiLlama-Adapters},
  2024.
\newblock URL
  \url{https://github.com/DefiLlama/DefiLlama-Adapters/blob/main/projects/ethena/index.js}.

\bibitem[{DefiPulse}(2021)]{DefiPulse2021-ai}
{DefiPulse}.
\newblock {Defi Pulse}.
\newblock \url{https://www.defipulse.com/blog}, 2021.
\newblock Accessed: 2024-6-22.

\bibitem[Dileo et~al.(2024)Dileo, Zignani, and Gaito]{Dileo2024-ep}
Manuel Dileo, Matteo Zignani, and Sabrina Gaito.
\newblock {Temporal graph learning for dynamic link prediction with text in
  online social networks}.
\newblock \emph{Machine Learning}, 113\penalty0 (4):\penalty0 2207--2226, April
  2024.
\newblock ISSN 0885-6125,1573-0565.
\newblock \doi{10.1007/s10994-023-06475-x}.
\newblock URL
  \url{https://link.springer.com/article/10.1007/s10994-023-06475-x}.

\bibitem[Doerr et~al.(2025)Doerr, Minot, Pandl, and
  Rossi]{Doerr2025-defi-lending}
J~F Doerr, J~Minot, K~Pandl, and P~Rossi.
\newblock Locked in, levered up: Risk, return, and ruin in {DeFi} lending.
\newblock \emph{Journal of International Money and Finance}, 151, 2025.
\newblock \doi{10.1016/j.jimonfin.2024.103234}.

\bibitem[Ester et~al.(1996)Ester, Kriegel, Sander, and Xu]{Ester1996-ou}
M~Ester, H~Kriegel, J~Sander, and Xiaowei Xu.
\newblock {A density-based algorithm for discovering clusters in large spatial
  databases with noise}.
\newblock \emph{Knowledge Discovery and Data Mining}, pages 226--231, August
  1996.
\newblock URL
  \url{https://cdn.aaai.org/KDD/1996/KDD96-037.pdf?source=post_page---------------------------}.

\bibitem[{European Systemic Risk Board}(2023)]{esrb2023crypto_defi}
{European Systemic Risk Board}.
\newblock Crypto-assets and decentralised finance: systemic implications and
  policy options.
\newblock Technical report, European Systemic Risk Board, 2023.
\newblock URL
  \url{https://www.esrb.europa.eu/pub/pdf/reports/esrb.cryptoassets~1b42e6b9f5.en.pdf}.

\bibitem[{Financial Stability Board}(2023)]{fsb2023defi}
{Financial Stability Board}.
\newblock The financial stability risks of decentralised finance.
\newblock Technical report, Financial Stability Board, February 2023.
\newblock URL
  \url{https://www.fsb.org/2023/02/the-financial-stability-risks-of-decentralised-finance/}.

\bibitem[Frazzini and Lamont(2005)]{Frazzini2005-iw}
Andrea Frazzini and Owen~A Lamont.
\newblock {Dumb Money: Mutual Fund Flows and the Cross-Section of Stock
  Returns}.
\newblock \emph{Social Science Research Network}, August 2005.
\newblock URL \url{https://papers.ssrn.com/abstract=776014}.

\bibitem[{International Standard
  Organization}(2021)]{International-Standard-Organization2021-xw}
{International Standard Organization}.
\newblock {ISO 10962:{2021}}, 2021.
\newblock URL \url{https://www.iso.org/standard/81140.html}.

\bibitem[Kitzler et~al.(2022)Kitzler, Victor, Saggese, and
  Haslhofer]{kitzler2022defi_compositions}
Stefan Kitzler, Friedhelm Victor, Pietro Saggese, and Bernhard Haslhofer.
\newblock Disentangling decentralized finance (defi) compositions.
\newblock \emph{ACM Transactions on the Web}, 16\penalty0 (4):\penalty0 1--36,
  2022.
\newblock \doi{10.1145/3532857}.

\bibitem[Lee et~al.(2023)Lee, Naifar, and Shahbaz]{lee2023terra_luna_case}
Junghoon Lee, Nader Naifar, and Muhammad Shahbaz.
\newblock Anatomy of a stablecoin’s failure: The terra-luna case.
\newblock \emph{Finance Research Letters}, 54:\penalty0 103883, 2023.
\newblock \doi{10.1016/j.frl.2023.103883}.

\bibitem[Li et~al.(2024)Li, Li, Zhang, and Li]{Li2024-ha}
Wenkai Li, Xiaoqi Li, Yuqing Zhang, and Zongwei Li.
\newblock {DeFiTail: DeFi Protocol Inspection through Cross-Contract Execution
  Analysis}.
\newblock In \emph{{Companion Proceedings of the ACM on Web Conference 2024}},
  pages 786--789, New York, NY, USA, May 2024. ACM.
\newblock ISBN 9798400701726.
\newblock \doi{10.1145/3589335.3651488}.
\newblock URL
  \url{https://dl.acm.org/doi/abs/10.1145/3589335.3651488?casa_token=uiz1Na13B4cAAAAA:ai_SpNcyMNrDb6i9eoxnxjKev_xoVU7TW-D81VJuvjfsJqj4vP_x4VVJR8HNrMHAx7qjMDHFsxOM}.

\bibitem[Luo et~al.(2024)Luo, Feng, Xu, and Tasca]{Luo2024-kj}
Yichen Luo, Yebo Feng, Jiahua Xu, and Paolo Tasca.
\newblock {Piercing the veil of TVL: DeFi reappraised}.
\newblock \emph{arXiv [q-fin.GN]}, April 2024.
\newblock URL \url{http://arxiv.org/abs/2404.11745}.

\bibitem[Lyons and Viswanath-Natraj(2020)]{Lyons2020-ie}
Richard~K Lyons and Ganesh Viswanath-Natraj.
\newblock {What keeps stablecoins stable?}
\newblock \emph{SSRN Electronic Journal}, 2020.
\newblock ISSN 1556-5068.
\newblock \doi{10.2139/ssrn.3597868}.
\newblock URL
  \url{https://www.jbs.cam.ac.uk/wp-content/uploads/2020/08/2020-conference-paper-lyons-viswanath-natraj.pdf}.

\bibitem[Ma et~al.(2025)Ma, Zhu, Liu, Xie, and Li]{Ma2024}
Wei Ma, Chenguang Zhu, Ye~Liu, Xiaofei Xie, and Yi~Li.
\newblock A comprehensive study of governance issues in decentralized finance
  applications.
\newblock \emph{ACM Transactions on Software Engineering and Methodology},
  34\penalty0 (7), 2025.
\newblock \doi{10.1145/3717062}.

\bibitem[Maaten and Hinton(2008)]{Maaten2008-ul}
L~Maaten and Geoffrey~E Hinton.
\newblock {Visualizing Data using t-{SNE}}.
\newblock \emph{Journal of Machine Learning Research}, 9:\penalty0 2579--2605,
  2008.
\newblock URL
  \url{https://www.jmlr.org/papers/volume9/vandermaaten08a/vandermaaten08a.pdf?fbcl}.

\bibitem[McLeay et~al.(2014)McLeay, Radia, and Thomas]{McLeay2014-fo}
M~McLeay, Amar Radia, and Ryland Thomas.
\newblock {Money creation in the modern economy}.
\newblock \emph{ERN: Banking \& Monetary Policy (Topic)}, March 2014.
\newblock URL
  \url{https://www.bankofengland.co.uk/-/media/boe/files/quarterly-bulletin/2014/money-creation-in-the-modern-economy}.

\bibitem[Metelski and Sobieraj(2022)]{Metelski2022-kf}
D~Metelski and J~Sobieraj.
\newblock {Decentralized finance (DeFi) projects: A study of key performance
  indicators in terms of DeFi protocols’ valuations}.
\newblock \emph{International Journal of Financial Studies}, November 2022.
\newblock ISSN 2227-7072.
\newblock \doi{10.3390/ijfs10040108}.
\newblock URL \url{https://www.mdpi.com/2227-7072/10/4/108}.

\bibitem[Nakamoto(2008)]{Nakamoto2008-cq}
Satoshi Nakamoto.
\newblock {Bitcoin whitepaper}.
\newblock \emph{URL: Https://bitcoin. Org/Bitcoin. PDF-(: 17. 07. 2019)}, 2008.
\newblock URL
  \url{https://huobi-1253283450.cos.ap-beijing.myqcloud.com/1543476765952_IgOT7VVGO4Vr3QUjymBa.pdf}.

\bibitem[Qaiser and Ali(2018)]{Qaiser2018-bn}
Shahzad Qaiser and R~Ali.
\newblock {Text mining: Use of TF-IDF to examine the relevance of words to
  documents}.
\newblock \emph{International Journal of Computer Applications}, July 2018.
\newblock ISSN 0975-8887.
\newblock \doi{10.5120/IJCA2018917395}.
\newblock URL
  \url{https://www.researchgate.net/profile/Shahzad-Qaiser/publication/326425709_Text_Mining_Use_of_TF-IDF_to_Examine_the_Relevance_of_Words_to_Documents/links/5b4cd57fa6fdcc8dae245aa3/Text-Mining-Use-of-TF-IDF-to-Examine-the-Relevance-of-Words-to-Documents.pdf}.

\bibitem[Qin et~al.(2021)Qin, Zhou, Gamito, Jovanovic, and
  Gervais]{Qin2021-liquidations}
Kaihua Qin, Liyi Zhou, Pablo Gamito, Philipp Jovanovic, and Arthur Gervais.
\newblock An empirical study of {DeFi} liquidations: Incentives, risks, and
  instabilities.
\newblock \emph{arXiv}, 2021.
\newblock Proceedings of the 21st ACM Internet Measurement Conference (IMC
  '21).

\bibitem[Ramos(2003)]{Ramos2003-cf}
J~E Ramos.
\newblock {Using TF-IDF to determine word relevance in document queries}.
\newblock \emph{Proceedings of the First Instructional Conference on Machine
  Learning}, 242\penalty0 (1):\penalty0 29--48, 2003.
\newblock URL
  \url{https://citeseerx.ist.psu.edu/document?repid=rep1&type=pdf&doi=b3bf6373ff41a115197cb5b30e57830c16130c2c}.

\bibitem[Rossi et~al.(2020)Rossi, Chamberlain, Frasca, Eynard, Monti, and
  Bronstein]{rossi2020tgn}
Emanuele Rossi, Benjamin Chamberlain, Fabrizio Frasca, Davide Eynard, Federico
  Monti, and Michael Bronstein.
\newblock Temporal graph networks for deep learning on dynamic graphs.
\newblock In \emph{ICML 2020 Workshop on Graph Representation Learning}, 2020.
\newblock URL \url{https://arxiv.org/abs/2006.10637}.

\bibitem[Saggese et~al.(2025)Saggese, Fr{\"o}wis, Kitzler, Haslhofer, and
  Auer]{saggese2025towards}
Pietro Saggese, Michael Fr{\"o}wis, Stefan Kitzler, Bernhard Haslhofer, and
  Raphael Auer.
\newblock Towards verifiability of total value locked (tvl) in decentralized
  finance.
\newblock In \emph{2025 IEEE International Conference on Blockchain and
  Cryptocurrency (ICBC)}, pages 1--9. IEEE, 2025.

\bibitem[Sch{\"a}r(2021)]{Schar2021-defi}
Fabian Sch{\"a}r.
\newblock Decentralized finance: On blockchain- and smart contract-based
  financial markets.
\newblock \emph{Federal Reserve Bank of St. Louis Review}, 103\penalty0
  (2):\penalty0 153--174, 2021.
\newblock \doi{10.20955/r.103.153-74}.
\newblock URL
  \url{https://research.stlouisfed.org/publications/review/2021/04/15/decentralized-finance-on-blockchain-and-smart-contract-based-financial-markets}.

\bibitem[Squartini et~al.(2013)Squartini, van Lelyveld, and
  Garlaschelli]{Squartini2013-qu}
Tiziano Squartini, Iman van Lelyveld, and Diego Garlaschelli.
\newblock {Early-warning signals of topological collapse in interbank
  networks}.
\newblock \emph{arXiv [physics.soc-ph]}, February 2013.
\newblock \doi{10.1038/srep03357}.
\newblock URL \url{http://dx.doi.org/10.1038/srep03357}.

\bibitem[Stepanova and Eriņš(2021)]{Stepanova2021-uo}
Viktorija Stepanova and Ingars Eriņš.
\newblock {Review of decentralized Finance applications and their Total Value
  Locked}.
\newblock \emph{TEM Journal}, pages 327--333, February 2021.
\newblock ISSN 2217-8309,2217-8333.
\newblock \doi{10.18421/tem101-41}.
\newblock URL \url{https://www.ceeol.com/search/article-detail?id=935904}.

\bibitem[Sun et~al.(2025)Sun, Ma, Nie, and Liu]{Sun2025}
Dianxiang Sun, Wei Ma, Liming Nie, and Yang Liu.
\newblock {SoK}: A taxonomic analysis of {DeFi} rug pulls: Types, dataset, and
  tool assessment.
\newblock \emph{Proceedings of the ACM on Software Engineering}, 2\penalty0
  (ISSTA):\penalty0 550--572, 2025.
\newblock \doi{10.1145/3728900}.

\bibitem[Suzuki et~al.(2025)Suzuki, Pa, Nguyen, and Yoshioka]{Suzuki2025}
Iori Suzuki, Yin Minn~Pa Pa, Anh Thi~Van Nguyen, and Katsunari Yoshioka.
\newblock {DeFiIntel}: A dataset bridging on-chain and off-chain data for
  {DeFi} token scam investigation.
\newblock In \emph{Proceedings of the 7th Workshop on Measurements, Attacks,
  and Defenses for the Web ({MADWeb} 2025)}, San Diego, CA, USA, 2025. Internet
  Society.
\newblock \doi{10.14722/madweb.2025.23019}.

\bibitem[Weingärtner et~al.(2023)Weingärtner, Fasser, Reis Sá~da Costa, and
  Farkas]{Weingartner2023-wx}
Tim Weingärtner, Fabian Fasser, Pedro Reis Sá~da Costa, and Walter Farkas.
\newblock {Deciphering DeFi: A comprehensive analysis and visualization of
  risks in decentralized finance}.
\newblock \emph{Journal of Risk and Financial Management}, 16\penalty0
  (10):\penalty0 454, October 2023.
\newblock ISSN 1911-8066,1911-8074.
\newblock \doi{10.3390/jrfm16100454}.
\newblock URL \url{https://www.mdpi.com/1911-8074/16/10/454}.

\bibitem[Werner et~al.(2022)Werner, Perez, Gudgeon, Klages-Mundt, Harz, and
  Knottenbelt]{Werner2022-ce}
Sam Werner, Daniel Perez, Lewis Gudgeon, Ariah Klages-Mundt, Dominik Harz, and
  William Knottenbelt.
\newblock {SoK: Decentralized Finance (DeFi)}.
\newblock In \emph{{Proceedings of the 4th ACM Conference on Advances in
  Financial Technologies}}, pages 30--46, New York, NY, USA, September 2022.
  ACM.
\newblock ISBN 9781450398619.
\newblock \doi{10.1145/3558535.3559780}.
\newblock URL \url{https://dl.acm.org/doi/abs/10.1145/3558535.3559780}.

\bibitem[Xu and Vadgama(2022)]{Xu2022-ni}
Jiahua Xu and Nikhil Vadgama.
\newblock {From Banks to DeFi: the Evolution of the Lending Market}.
\newblock In Nikhil Vadgama, Jiahua Xu, and Paolo Tasca, editors,
  \emph{{Enabling the Internet of Value: How Blockchain Connects Global
  Businesses}}, pages 53--66. Springer International Publishing, Cham, 2022.
\newblock ISBN 9783030781842.
\newblock \doi{10.1007/978-3-030-78184-2\_6}.
\newblock URL \url{https://doi.org/10.1007/978-3-030-78184-2_6}.

\bibitem[Xu et~al.(2021)Xu, Paruch, Cousaert, and Feng]{Xu2021-bh}
Jiahua Xu, Krzysztof Paruch, Simon Cousaert, and Yebo Feng.
\newblock {SoK: Decentralized Exchanges (DEX) with Automated Market Maker (AMM)
  Protocols}.
\newblock \emph{arXiv [q-fin.TR]}, \penalty0 (April), March 2021.
\newblock URL \url{http://arxiv.org/abs/2103.12732}.

\bibitem[Xu and Xu(2023)]{Xu2023-pf}
Teng~Andrea Xu and Jiahua Xu.
\newblock {A Short Survey on Business Models of Decentralized Finance (DeFi)
  Protocols: CoDecFin, DeFi, Voting, WTSC, Grenada, May 6, 2022, Revised
  Selected Papers}.
\newblock In \emph{{Financial Cryptography and Data Security. FC 2022
  International Workshops}}, volume 13412 of \emph{Lecture Notes in Computer
  Science}, pages 197--206, Cham, 2023. Springer International Publishing.
\newblock ISBN 9783031324147.
\newblock \doi{10.1007/978-3-031-32415-4\_13}.
\newblock URL
  \url{https://link.springer.com/chapter/10.1007/978-3-031-32415-4_13}.

\bibitem[You et~al.(2022)You, Du, and Leskovec]{You2022-oy}
Jiaxuan You, Tianyu Du, and Jure Leskovec.
\newblock {ROLAND: Graph learning framework for dynamic graphs}.
\newblock \emph{arXiv [cs.LG]}, August 2022.
\newblock URL \url{http://arxiv.org/abs/2208.07239}.

\bibitem[Zetzsche et~al.(2020)Zetzsche, Arner, and Buckley]{Zetzsche2020-pv}
D~A Zetzsche, D~W Arner, and R~P Buckley.
\newblock {Decentralized finance (defi)}.
\newblock \emph{Journal of Financial Regulation}, 6:\penalty0 172--203, 2020.
\newblock \doi{10.1093/jfr/fjaa010/37064506/fjaa010}.
\newblock URL
  \url{https://papers.ssrn.com/sol3/papers.cfm?abstract_id=3682212}.

\bibitem[Zhang(2023)]{zhang2023machine}
Luyao Zhang.
\newblock Machine learning for blockchain: Literature review and open research
  questions.
\newblock In \emph{NeurIPS 2023 AI for Science Workshop}, 2023.

\bibitem[Zhang et~al.(2023)Zhang, Luo, Lu, and He]{Zhang2023-dt}
Zhen Zhang, B~Luo, Shengliang Lu, and Bingsheng He.
\newblock {Live Graph Lab: Towards open, dynamic and real transaction graphs
  with {NFT}}.
\newblock \emph{Neural Information Processing Systems},
  abs/2310.11709:\penalty0 18769--18793, October 2023.
\newblock \doi{10.48550/arXiv.2310.11709}.
\newblock URL
  \url{https://proceedings.neurips.cc/paper_files/paper/2023/hash/3be31c1a2fdcb7b748c53c3f4cb0e9d2-Abstract-Datasets_and_Benchmarks.html}.

\bibitem[Znaidi et~al.(2023)Znaidi, Sia, Ronquist, Rajapakse, Jonckheere, and
  Bogdan]{Znaidi2023-jd}
Mohamed~Ridha Znaidi, Jayson Sia, Scott Ronquist, Indika Rajapakse, Edmond
  Jonckheere, and Paul Bogdan.
\newblock {A unified approach of detecting phase transition in time-varying
  complex networks}.
\newblock \emph{Scientific Reports}, 13\penalty0 (1):\penalty0 17948, October
  2023.
\newblock ISSN 2045-2322,2045-2322.
\newblock \doi{10.1038/s41598-023-44791-3}.
\newblock URL \url{https://www.nature.com/articles/s41598-023-44791-3}.

\end{thebibliography}

\FloatBarrier

\end{document}